\documentclass[11pt]{article}
\usepackage{appendix}
\usepackage[figuresright]{rotating}
\usepackage{amsmath,amssymb,amscd,amsthm}
\usepackage{graphicx}
\usepackage{dcolumn}
\usepackage{bm}
\usepackage{ifthen}
\usepackage{rh_defs_21}

\usepackage{hyperref}
\hypersetup{
    bookmarks=true,         
    unicode=false,          
    pdftoolbar=true,        
    pdfmenubar=true,        
    pdffitwindow=false,     
    pdfstartview={FitH},    
    pdfauthor={Reinhard Heckel},     
    pdfsubject={Subject},   
    pdfcreator={Reinhard Heckel},   
    pdfproducer={Producer}, 
    pdfnewwindow=true,      
    colorlinks=true,       
    linkcolor=red,          
    citecolor=green,        
    filecolor=magenta,      
    urlcolor=cyan           
}

\usepackage{filecontents}

\usepackage[
backend=bibtex,
url=false,isbn=false,doi=false,
date=year,
hyperref=auto,
style=alphabetic,
natbib=true,
sorting=nty,
firstinits=true,
maxnames=2, maxbibnames=10,
]{biblatex}

\bibliography{./bibliography.bib} 

\usepackage{comment}

\usepackage{tikz}
\usepackage{pgfplots}
\usepgfplotslibrary{groupplots} 
\usetikzlibrary{pgfplots.groupplots}
\usetikzlibrary{matrix,arrows,decorations.pathmorphing}
\usepackage{pgfplots,pgfplotstable}
\usepgfplotslibrary{fillbetween}
\pgfplotsset{compat=1.10}

\usetikzlibrary{matrix,arrows,decorations.pathmorphing}

\usetikzlibrary{external}

\usepackage{amsmath,amssymb,ifthen,color,framed,bm}

\newtheorem{lemma}{Lemma}

\newtheorem{theorem}{Theorem}

\newtheorem{proposition}{Proposition}

\newcommand\pdim{p}
\newcommand\Jac{\mJ}
\newcommand\risk{R}
\newcommand\emprisk{\hat R}
\newcommand\relu{\mathrm{relu}}
\newcommand\iter{t}
\newcommand\loss{\ell}
\newcommand\ind[1]{1\{#1\}}
\newcommand\stepsize{\eta}

\begin{document}

\begin{center}

{\bf{\LARGE{
Provable Continual Learning via Sketched Jacobian Approximations
}}}

\vspace*{.2in}

{\large{
\begin{tabular}{cccc}
Reinhard Heckel$^{\ast,\dagger}$
\end{tabular}
}}

\vspace*{.05in}

\begin{tabular}{c}
$^\ast$Dept. of Electrical and Computer Engineering, Technical University of Munich\\
$^\dagger$Dept. of Electrical and Computer Engineering, Rice University
\end{tabular}

\vspace*{.1in}

\today

\vspace*{.1in}

\end{center}


\begin{abstract}
An important problem in machine learning is the ability to learn tasks in a sequential manner. If trained with standard first-order methods most models forget previously learned tasks when trained on a new task, which is often referred to as catastrophic forgetting.
A popular approach to overcome forgetting is to regularize the loss function by penalizing models that perform poorly on previous tasks. For example, elastic weight consolidation (EWC) regularizes with a quadratic form involving a diagonal matrix build based on past data. 
While EWC works very well for some setups, we show that, even under otherwise ideal conditions, it can provably suffer catastrophic forgetting if the diagonal matrix is a poor approximation of the Hessian matrix of previous tasks. 
We propose a simple approach to overcome this: Regularizing training of a new task with sketches of the Jacobian matrix of past data. This provably enables overcoming catastrophic forgetting for linear models and for wide neural networks, at the cost of memory. 
The overarching goal of this paper is to provided insights on when regularization-based continual learning algorithms work and under what memory costs.
\end{abstract}


\section{Introduction}
Consider the problem of learning a number of tasks sequentially. 
Even if a neural network has the ability to perform well on all tasks simultaneously, if trained sequentially on different tasks, it tends to perform well only on the task it has most recently been trained on~\citep{french_CatastrophicForgettingConnectionist_1999,robins_CatastrophicForgettingRehearsal_1995,goodfellow_EmpiricalInvestigationCatastrophic_2015}. 
This is known as catastrophic forgetting.

Continual learning algorithms address forgetting and aim to enable sequential learning of several tasks.
Continual learning algorithms are often categorized into 
replay methods, regularization based methods, and approaches that modify a model directly by freezing/masking parts of the model or adding new parts to the model, see for example~\citep{rusu_ProgressiveNeuralNetworks_2016}. 

Replay methods~\citep{robins_CatastrophicForgettingRehearsal_1995,rebuffi_ICaRLIncrementalClassifier_2017,shin_ContinualLearningDeep_2017,li_LearningForgetting_2018} store past task data, or a generative model, and reuse past task data or pseudo-labels generated by the stored generative model when training a new task. 

Regularization based methods penalizes models with parameters far from those found important for previous tasks when training a new task. 
Two widely used regularization based methods are elastic weight consolidation (EWC)~\citep{kirkpatrick_OvercomingCatastrophicForgetting_2017} and synaptic intelligence (SI)~\citep{zenke_ContinualLearningSynaptic_2017}. Both 
 penalize the change of individual model coefficients deemed important for previous tasks via a quadratic penalty associated with each weight (or model parameter). 
This penalty ignores interactions between coefficients. Subsequent works incorporated quadratic penalties on the weights that take interactions between the weights into account, and/or provided generalizations of EWC~\citep{ritter_OnlineStructuredLaplace_2018,liu_RotateYourNetworks_2018,chaudhry_RiemannianWalkIncremental_2018,schwarz_ProgressCompressScalable_2018,pan_ContinualDeepLearning_2021}. This generally tends to improve performance. 
Versions of regularization-based methods are also referred to as online-Laplace algorithms, since they can be understood as applying a Laplace-approximation to the posterior.

Regularization based continual learning algorithms are appealing for its simplicity, but---like continual learning algorithms in general---are not well understood theoretically. It is important to study continual learning algorithms theoretically to understand when they work and when they fail. For example,  methods perform well on a given continual learning task, but fail on another. 
Taking the EWC algorithm as an example, it performs almost optimally on the MNIST permutation problem (see~Fig.~3B in \citep{kirkpatrick_OvercomingCatastrophicForgetting_2017}), but fails for learning pairs of different digits sequentially (see.~Fig.~2a in \citep{kemker_MeasuringCatastrophicForgetting_2018}). 

Motivated this discrepancy, the goal of this paper is to improve the understanding of regularization based continual learning methods by studying a family of algorithms that rely on approximating the loss functions pertaining to different tasks with random projections.  
The intuition is, that the outer product of the Jacobian is an approximation to the Hessian, and approximate Jacobians can efficiently approximate the Hessian. The family of algorithms uses the Hessian approximation in a quadratic penalty which approximates the loss function pertaining to past data. 
The EWC algorithm and adding an L2-penalty penalizing the move of coefficients are special cases. 

Our contributions are as follows:
\begin{itemize}
%
\item 
Our main result is to show that a regularization based continual learning algorithm trained with $\ell_2$-loss that works with sketched Jacobians provably enables continual learning, both for linear models and for wide neural networks. 
%
\item We conduct experiments on the MNIST permutation problem and the incremental MNIST task. 
The results show that working with a coarse sketch of the Jacobian gives significant improvements over the EWC algorithm, albeit at the cost of storing significantly more data.
\item The EWC algorithm and even importance/L2-regularization perform well for continual learning on the MNIST permutation problem. This is surprising, because the penalty used in the EWC algorithm is not a good approximation of the loss function of past data. To understand this, we study a model of the permutation problem theoretically, and show that both EWC and constant importance can work optimally on that task. 
We complement this result with a statement showing that EWC and importance/L2-regularization can provably fail as well. 
\end{itemize}
Those results contribute to an understanding of when we expect particular regularization based continual learning algorithms to work well and when not.

\subsection{Related work}
There are relatively few theoretical works on continual learning, compared to the vast literature on algorithms for continual learning and corresponding empirical results. Some recent theoretical developments include~\citep{yin_OptimizationGeneralizationRegularizationBased_2020,bennani_GeneralisationGuaranteesContinual_2020b,alquier_RegretBoundsLifelong_2017,doan_TheoreticalAnalysisCatastrophic_2021}. 
\citet{yin_OptimizationGeneralizationRegularizationBased_2020} studies optimization and generalization aspects of regularization based algorithms involving the Hessian. Our work differs in that we study a different family of algorithms (including random projections to approximate the Jacobian), and different models of data. 

Our work builds on the popular idea of using (an approximation of) the second-order Taylor expansion to approximate the loss of past data. This idea was proposed as early as in~\citep{ruvolo_ELLAEfficientLifelong_2013}, and of course the regularization based methods mentioned earlier can be viewed as being based on this idea. 
Finally, the very recent paper~\citep{li_LifelongLearningSketched_2021} also proposed to use random projections for sketching a regularizer involving the Jacobian, and provided experiments showing that this approach works well. Our works are very complementary, in that we provide explicit theoretical results showing that sketching the Jacobian enables provably continual learning for both linear models and for two-layer neural networks. In addition, we also provide positive and negative results for EWC. 

\section{Problem statement}

We consider the problem of learning a series of regression or classification tasks $A,B, \ldots$ sequentially. 
For each task $T$, we are given a set of training examples $\{(\vx_{T,1},y_{T,1}),\ldots, (\vx_{T,n},y_{T,n})\} \in \reals^d \times \setY$ drawn iid from an unknown distribution $P_T$ pertaining to task $T$. Here $\setY = \reals$ for a regression task, and $\setY = \{1,\ldots,Q\}$ for a classification task. 
Our goal is to train a single model $f_\vtheta \colon \reals^d \to \setY$ sequentially on the training sets of tasks $A,B,\ldots$, so that after training on task $T$, the model performs well \emph{on all past tasks}.

More specifically, after training on tasks $A,B,\ldots,T$, the model should perform well in predicting the response $y$ based on the feature vector $\vx$, with the unseen example $(\vx,y)$ drawn with equal probability from one of the distributions of tasks $A,B,\ldots, T$. Throughout, we assume that the method does not know at test time on which task it is evaluated. This is sometimes called single-head evaluation, as opposed to multi-head evaluation, where the task ID is known~\citep{chaudhry_RiemannianWalkIncremental_2018}, and is considered the more challenging and more practical evaluation mode or setup. 

\section{A family of regularization based continual learning algorithms}

We start with introducing a family of regularization-based continual learning algorithms that rely on approximations of the loss functions pertaining to different tasks. This family incorporates the EWC algorithm and is closely related to other regularization-based methods including~\citep{ritter_OnlineStructuredLaplace_2018,liu_RotateYourNetworks_2018,chaudhry_RiemannianWalkIncremental_2018,schwarz_ProgressCompressScalable_2018} in that it approximates the loss associated with past tasks.


The family of algorithms is parameterized by an approximation of the Jacobian of the model $f_\vtheta \colon \reals^d \to \setY$ with model parameter $\vtheta \in \reals^{\pdim}$ applied to the data of a task. Specifically, denote by $\mK_T \in \reals^{s\times \pdim }$ an approximation of the Jacobian $\Jac_T \in \reals^{n\times \pdim }$ of the predictions of the data of task $T$:
\[
\vf_T(\vtheta) = \transp{[f_{\vtheta}(\vx_{T,1}), \ldots, f_{\vtheta}(\vx_{T,n})]}.
\]
Here, $n$ is the number of training examples. 
The Jacobian $\mJ_T$ contains the gradients $\nabla_\vtheta f_\vtheta(\vx_{T,i}), i=1,\ldots,n$ as rows. 
The gradients have dimension $\pdim$ for a regression problem, and dimension ${(Q-1) \times \pdim}$ for a classification problem with $Q$ classes.


We train a model $f_\vtheta$ with a quadratic loss on the data from tasks $A,B,\ldots,T$ as follows.
First, we minimize the training loss of task $A$:
\begin{align}
\label{eq:lossA}
\mc L_{A}(\vtheta) = \frac{1}{2} \sum_{i=1}^n (f_\vtheta(\vx_{A,i}) - y_{A,i} )^2.
\end{align}
Let $\vtheta_A$ be a minimizer of this loss function. 
Next, we learn task $B$ by minimizing the loss 
\begin{align}
\label{eq:lossAB}
\mc L_{AB}(\vtheta) 
= 
\mc L_{B}(\vtheta) + \frac{\lambda}{2} \transp{(\vtheta - \vtheta_A)} \transp{\mK}_A\mK_A(\vtheta - \vtheta_A),
\end{align}
where $\mK_A \in \reals^{n \times \pdim}$ is the approximation of the Jacobian of the function 
$\vf_A(\vtheta) = \transp{[f_{\vtheta}(\vx_{A,1}), \ldots, f_{\vtheta}(\vx_{A,n})]}$ at $\vtheta = \vtheta_A$, and $\lambda \geq 0$ is a regularization parameter. 
In order to learn a third task $C$ we minimize the function
\begin{align*}
\mc L_{ABC}(\vtheta)
&= 
\mc L_{C}(\vtheta) + \frac{\lambda}{2} \transp{(\vtheta - \vtheta_{AB})} (\transp{\mK}_A\mK_A + \transp{\mK}_B\mK_B)(\vtheta - \vtheta_{AB}),
\end{align*}
where $\vtheta_{AB}$ is a minimum of $\mc L_{AB}(\vtheta)$ and $\mK_{B}$ is the approximate Jacobian of the predictions $\vf_B(\vtheta) = \transp{[f_{\vtheta}(\vx_{B,1}), \ldots, f_{\vtheta}(\vx_{B,n})]}$ at $\vtheta = \vtheta_{AB}$. 
The algorithm proceeds by learning further tasks $D, E,\ldots$ analogously. 


In the coming sections, we discuss four variants. All variants require at most computation of the Jacobian. The Jacobian is easy to compute for most popular machine learning models, because it simply requires computation of the gradients on the training examples, and first order methods for optimization already compute those in each epoch. We state the memory requirement for $K$ tasks.
\begin{enumerate}
\item[i)] {\bf Regularization with Original Jacobian:} Take the approximation of the Jacobian as the Jacobian of the predictions $\vf_T(\vtheta)$ at the appropriate $\vtheta$, i.e., $\mK_T = \mJ_T$.
Memory requirement: $p(1+ Kn)$.
\item[ii)] {\bf Regularization with Sketched Jacobian (RSJ):} Take the approximation of the Jacobian as a random sketch of the Jacobian of the predictions $\vf_T(\vtheta)$ at the appropriate $\vtheta$, obtained by left-multiplying the Jacobian with a Gaussian random projection matrix $\mS_T \in \reals^{s\times n}$, with iid $\mc N(0,1/s)$ entries, i.e., $\mK_T = \mS_T \mJ_T \in \reals^{s\times \pdim}$. 
Memory requirement: $p(1+ Ks)$.
Note that this algorithm is indexed by $s$, therefore we refer to it as RSJ-$s$ in the following (e.g., RSJ-$50$ is RSJ with $s=50$). 
\item[iii)] {\bf EWC:} Take the approximation of the Jacobian $\mK_T \in \reals^{p\times \pdim}$ as the square-root of the diagonal of the outer product of the Jacobians $\transp{\mJ}_T \mJ_T$. This algorithm is a variant of the EWC algorithm.  Specifically, it corresponds to ``online'' EWC where instead of the original Fisher matrix, the empirical Fisher matrix is used (see~\citep{kunstner_LimitationsEmpiricalFisher_2019} on the relations of the original and empirical Fisher matrices). 
Memory requirement: $2p$.
\item[iv)] {\bf L2:} Take the approximation of the Jacobian $\mK_T  \in \reals^{\pdim\times \pdim}$ as the identity matrix. This amounts to simple constant importance/L2-regularization penalizing the movement of coefficients from one task to the other. Memory requirement: $p$.
\end{enumerate}

The intuition behind this family of algorithms is as follows. 
First, consider the variant that takes the exact Jacobian as the matrix $\mK_A$. 
If the model $f_\vtheta(\vx)$ is linear in the model parameter $\vtheta$ 
then, as shown in the next section, for $\lambda=1$, the algorithm performs optimal continual learning, since the model learned at each step $T$ is equivalent to the model learned when training on all data $A,B,\ldots,T$. 
The model is linear in the model parameter for all kernel methods, and is approximately linear for wide neural networks, as established by the recent theory on the neural tangent kernel~\citep{jacot_NeuralTangentKernel_2018,lee_DeepNeuralNetworks_2018a,du_GradientDescentProvably_2018b,oymak_ModerateOverparameterizationGlobal_2020}. 
For this linear setup, the importance matrix $\transp{\mJ}_A \mJ_A$ is equal to the Hessian matrix. 

While working with the original Jacobian gives a continual learning algorithm that provably succeeds for linear models, the corresponding algorithm is impractical for large commonly used models. For small, under-parameterized toy models working with the original Jacobian is a viable approach, but for large, over-parameterized models used in practice it is infeasible and impractical to store all the Jacobians, because in that case the memory requirement would be larger than that of storing the training data. 

The intuition behind regularization with the sketched Jacobian (RSJ algorithm) is that, if we manage to obtain accurate approximation of the Jacobian outer products $\transp{\mJ}_T \mJ_T$ for all tasks, then we expect the corresponding algorithm to behave similar to the optimal algorithm that works with the original Jacobian. 
In Section~\ref{sec:theorylinearmodel} and \ref{sec:NTK} we show that the sketched Jacobians can provably enable continual learning for linear models and wide neural networks. This comes at a price: larger values of $s$ give a better approximation at a higher memory cost.

The third variant, the EWC algorithm, can be viewed as taking an extreme approach to approximating the Jacobian outer product $\transp{\mJ}_T \mJ_T$ by simply taking its diagonal. 

\section{Empirical observations}
\label{sec:emp}
We start by evaluating the methods introduced in the previous section on two popular continual learning problems: The MNIST 
permutation problem 
and the incremental MNIST problem. 
We will make a number of empirical observations, and then explain those empirical observations with theoretical results in the remaining sections. All simulations were run on a single RTX 5000 GPU and are reproducible with the code in the supplement.

There are a variety of interesting new state-of-the-art approaches for continual learning. We do not compare to those, because our goal is not to establish a new state-of-the-art method but rather to understand regularization based continual learning algorithms better. However, we compare to ``training on all data'' as a reference point, which can be viewed as an upper bound for any continual learning algorithm.

\subsection{MNIST permutation Problem}

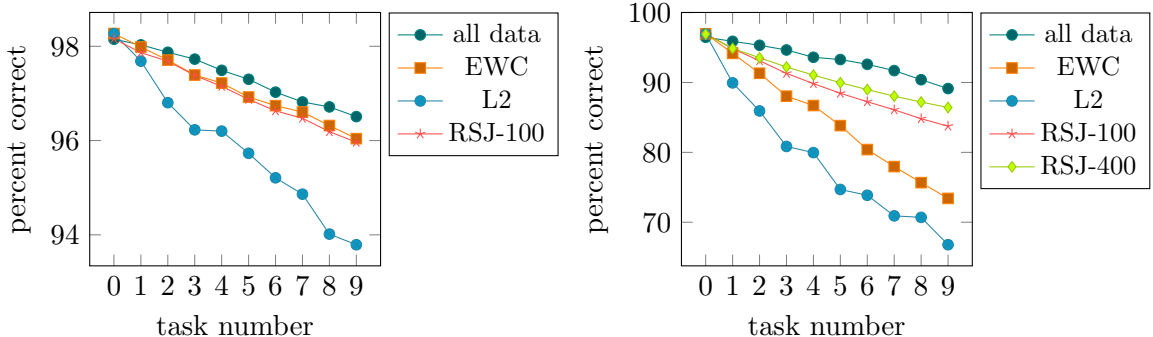
\begin{figure}[t]
\begin{center}
\begin{tikzpicture}
\begin{groupplot}[
legend style={at={(1.32,1)},anchor=north}, 
group style={group size=2 by 1, horizontal sep=4cm },
width=0.33\textwidth,height=0.3\textwidth,
xticklabel style={/pgf/number format/fixed, /pgf/number format/precision=3},
yticklabel style={/pgf/number format/fixed, /pgf/number format/precision=5},
]
\nextgroupplot[xlabel = {task number},ylabel={percent correct},xtick={0,...,9},
 cycle list name=exotic,]
  \addplot + [] table[x index=0,y index=2]{./fig/MNIST_permutation_2layernn.dat};
  \addlegendentry{\small all data};
  \addplot + [] table[x index=0,y index=1]{./fig/MNIST_permutation_2layernn.dat};
  \addlegendentry{\small EWC};
  \addplot + [] table[x index=0,y index=5]{./fig/MNIST_permutation_2layernn.dat};
  \addlegendentry{\small L2};
  
  \addplot + [] table[x index=0,y index=3]{./fig/MNIST_permutation_2layernn.dat};
  \addlegendentry{\small RSJ-$100$};

\nextgroupplot[xlabel = {task number},xtick={0,...,9},
 cycle list name=exotic,ylabel={percent correct},]

  \addplot + [] table[x index=0,y index=2]{./fig/MNIST_permutation_rand_feat.dat};
  \addlegendentry{\small all data};
  
  \addplot + [] table[x index=0,y index=1]{./fig/MNIST_permutation_rand_feat.dat};
  \addlegendentry{\small EWC};

  \addplot + [] table[x index=0,y index=5]{./fig/MNIST_permutation_rand_feat.dat};
  \addlegendentry{\small L2};
    
  \addplot + [] table[x index=0,y index=3]{./fig/MNIST_permutation_rand_feat.dat};
  \addlegendentry{\small RSJ-$100$};

  \addplot + [] table[x index=0,y index=4]{./fig/MNIST_permutation_rand_feat.dat};
  \addlegendentry{\small RSJ-$400$};

\end{groupplot}          
\end{tikzpicture}
\end{center}

\vspace{-0.5cm}
\caption{
\label{eq:permtask}
{\bf Left:} Sequential learning on the MNIST permutation problem for a two-layer fully connected relu network. 
All data refers to training on all the data of all the tasks.  
The experiment shows that on this task, if correctly tuned, all algorithms perform close to optimal, i.e., close to training sequentially on all data. 
{\bf Right:} 
Sequential learning on the MNIST permutation problem for relu-random feature model with $6\cdot784$ random features. 
For this model, which is much smaller than the two-layer model from the past experiment, EWC and L2 regularization do not work well, but the RSJ algorithm which uses a much better approximation of the Jacobian significantly improves performance.
\label{fig:permtask}
}
\end{figure}

The popular MNIST permutation problem~\citep{goodfellow_EmpiricalInvestigationCatastrophic_2015,kirkpatrick_OvercomingCatastrophicForgetting_2017} is as follows. 
Task A is the original MNIST digit classification problem,  and all remaining tasks are obtained by permuting the pixels of each image with a random permutation that is fixed for each task $B,C,\ldots$.
This is often called task-incremental learning. 
We apply the family of algorithms introduced in the previous section to continually learning $10$ of such tasks with a two-layer fully connected network with $500$ hidden nodes and relu-activation functions. 
Figure~\ref{eq:permtask} shows the results, including the baseline ``all data'', which means the network has been trained on all data from all tasks. This baseline is an upper bound on the performance of any continual learning algorithm.

We first observe that, perhaps surprisingly, {\bf EWC performs extremely well on permuted MNIST}:
Almost as well as training on all data from all tasks. 
We find this surprising, because the EWC algorithm works well here even though the Jacobian outer product is not well approximated with its diagonal. To see that the Jacobian outer product, $\transp{\mJ}_T \mJ_T$, cannot be well-approximated by its diagonal, note that the number of training examples is $n=60000$, while the number of model parameters is $p=397510$, therefore $\transp{\mJ}_T \mJ_T$ has rank at most $n$ while its diagonal has rank $p \gg b$. 

The experiment also shows that simple constant importance/L2-regularization performs well, it is only 2\%-less than optimal on $10$ tasks. Note that for L2-regularization to work well it is critical to scale the penalty of each set of parameters (first layer weights, first layer bias, second layer weights, second layer bias) appropriately (through hyperparameter optimization), which we have done here. 

In Section \ref{sec:EWCtheory} we explain theoretically why EWC (and even constant importance/L2-regularization) can perform so well in some situations. 

Also note that the random projection based algorithm RSJ-100 performs similar to EWC for this setup, but not better, because there is little room for improvement to training on all data. 

Next, we study the performance of the family of continual learning algorithms on a much smaller model, specifically on a random feature model, with $6 \cdot 784$ Gaussian random relu features (i.e., we fit the model $f_\vtheta(\vx) = \relu(\mTheta \vx) \vtheta$, where $\mTheta$ is a Gaussian random matrix). This model has much fewer parameters than the neural network considered earlier and is linear in the model parameter $\vtheta$ (but not in $\vx$). 
Figure~\ref{eq:permtask}, right panel, depicts the results. 

The results show that for this smaller model, the EWC and L2-regularization algorithms perform significantly worse than the random-projection based RSJ algorithm algorithm with a sufficiently large random projection dimension ($s=400$). The RSJ-400 algorithm performs almost on par with training on all data. 

An important observation from this experiment is that the 
{\bf the gap between learning on all data and the RP based algorithm increases from task to task}.
That is intuitively expected because the error due to the approximate Jacobians accumulates from task to task. We discuss this aspect in Section~\ref{sec:theorylinearmodel} theoretically. 

\subsection{Incremental MNIST problem}

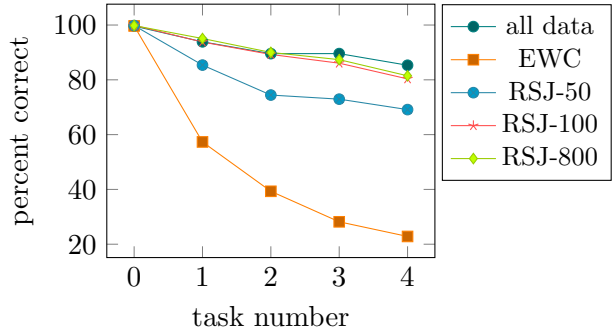
\begin{figure}[t]
\begin{center}
\begin{tikzpicture}
\begin{groupplot}[
ylabel={percent correct},
xtick={0,...,4},
 cycle list name=exotic,
legend style={at={(1.28,1)},anchor=north}, 
group style={group size=2 by 1, horizontal sep=1.2cm },
width=0.36\textwidth,height=0.3\textwidth, 
xticklabel style={/pgf/number format/fixed, /pgf/number format/precision=3},
yticklabel style={/pgf/number format/fixed, /pgf/number format/precision=5}
]
\nextgroupplot[xlabel = {task number}]
  \addplot + [] table[x index=0,y index=2]{./fig/MNIST_incremental_rand_feat.dat};
  \addlegendentry{\small all data};
  
  \addplot + [] table[x index=0,y index=1]{./fig/MNIST_incremental_rand_feat.dat};
  \addlegendentry{\small EWC};

  \addplot + [] table[x index=0,y index=3]{./fig/MNIST_incremental_rand_feat.dat};
  \addlegendentry{\small RSJ-$50$};
    \addplot + [] table[x index=0,y index=4]{./fig/MNIST_incremental_rand_feat.dat};
  \addlegendentry{\small RSJ-$100$};
  \addplot + [] table[x index=0,y index=5]{./fig/MNIST_incremental_rand_feat.dat};
  \addlegendentry{\small RSJ-$800$};

\end{groupplot}          
\end{tikzpicture}
\end{center}

\vspace{-0.5cm}
\caption{
\label{eq:permtaskrf}
Sequentially learning to classify the digits $\{0,1\},\{2,3\},\ldots,\{8,9\}$ with a relu-random feature model with $6 \cdot 784$ random features. 
For this problem, EWC does not work at all, but the RSJ algorithm with a sufficiently large random projection (i.e., value of $s$) works almost as well as training on all data.
}
\end{figure}

Finally, we study the problem of incrementally learning to classify digits. This is called class incremental learning by \citet{vandeven_ThreeScenariosContinual_2019,hsu_ReevaluatingContinualLearning_2019}). Task $A$ is to classify $\{0,1\}$, task $B$ to classify $\{2,3\}$, etc, until task $E$ which is to classify $\{8,9\}$. This problem and variants thereof are a popular continual learning baseline~\citep{kemker_MeasuringCatastrophicForgetting_2018,zenke_ContinualLearningSynaptic_2017}. 
Figure~\ref{eq:permtaskrf} shows the performance of the family of algorithms on the incremental learning task.
Note that after learning say tasks $A = \{0,1\}$ and $B=\{2,3\}$, the method is evaluated on both tasks simultaneously, i.e., on the test set containing all digits $\{0,1,2,3\}$. This assumes the task on which we test on is unknown, in contrast to the so called multi-head evaluation mode where the task is known. This evaluation mode is considered much harder, as mentioned earlier.

The results, depicted in Figure~\ref{eq:permtaskrf}, show that while EWC fails dramatically for this task (which is well known, cf.~Fig.~2a in \citep{kemker_MeasuringCatastrophicForgetting_2018}, the RSJ algorithm works almost as well as training on all the data, already for a projection dimension of $s=100$. 

\section{Guarantees for linear models}

\label{sec:theorylinearmodel}

We start by providing guarantees for the Jacobian regularization based learning algorithms (based on the full and sketched Jacobian) for linear models.
We consider models $f_\vtheta(\vx)$ that are linear in $\vtheta$, i.e., there exists a feature map $\psi \colon \reals^{d}\to \reals^{p}$, so that $f_{\vtheta}(\vx) = \innerprod{\psi(\vx)}{\vtheta}$. 
All kernel methods can be written in this form (although for some kernels this feature map is infinite dimensional). 

We start with an illustrative result that guarantees that if we work with the original Jacobian, then regularization based continual learning is provably correct:

\begin{proposition}
\label{eq:proplin}
Suppose that the model $f_{\vtheta}$ is linear in $\vtheta$, and consider the continual learning algorithm with the original Jacobian and with regularization parameter $\lambda=1$ trained with $\ell_2$-loss. Then the model learned with Jacobian regularization gives exactly the same result as training on the original data.
\end{proposition}

We hasten to add that it might be obvious to experts in continual learning that regularization based continual learning with the exact Jacobian provably succeeds for linear models; we nevertheless state this formally to put the results to come into context.

As mentioned before, for small, under-parameterized toy models working with the original Jacobian is a viable approach, but for large, over-parameterized models used in practice it is infeasible and impractical, because the memory requirement of storing the Jacobian is larger than that of storing the entire training data. 

We next discuss the RSJ algorithm that works with the sketched Jacobian, and for simplicity we focus on learning two tasks $A$ and $B$. Learning of task $A$ is straightforward and amounts to minimizing the least-squares loss on task $A$ (i.e., $\mc L_{A}(\vtheta)$). 
Learning task $B$ after having learned task $A$ amounts to solving the partially sketched least-squares problem:
\begin{align*}
\mc L_{AB}(\vtheta) 
&=
\mc L_{B}(\vtheta) + \frac{1}{2} \transp{(\vtheta - \vtheta_A)} \transp{\mJ}_A \transp{\mS} \mS \mJ_A(\vtheta - \vtheta_A).
\end{align*}
This can be viewed as a perturbed version of the least-squares problem
\begin{align}
\label{eq:lossall}
\tilde{\mc L}_{AB}(\vtheta)
=
\mc L_{A}(\vtheta) + \mc L_{B}(\vtheta)
=
\frac{1}{2} \norm[2]{\mJ \vtheta - \vy}^2,
\end{align}
corresponding to training on the entire data. 
To see this, note that if the random projection dimension is sufficiently large then the matrix $\transp{\mJ}_A \transp{\mS} \mS \mJ_A$ is a good approximation of the matrix $\transp{\mJ}_A \mJ_A$, which in turn means that the loss $\mc L_{AB}$ is a good approximation of the loss $\tilde{\mc L}_{AB}$.  
Here, we defined 
\begin{align*}
\mJ 
= 
\begin{bmatrix}
\mJ_A \\ \mJ_B
\end{bmatrix},
\quad
\vy
=
\begin{bmatrix}
\mJ_A \vtheta_A \\ \vy_B
\end{bmatrix}.
\end{align*}
Note that the sketch of the Jacobian induces a perturbation, and therefore the solution obtained by minimizing the loss $\mc L_{AB}(\vtheta)$ is in general different than minimizing the loss of the entire data. However, if the Jacobian of $\mJ_A$ is well approximated by a matrix of rank $r$, then we expect that a sketch of dimension $s = O(r)$ approximates the Jacobian well, and the estimate obtained by minimizing the sketched least-squares problem is expected to be close to the solution obtained by minimizing the original least-squares problem. 

The following result, proven in the supplement, formalizes this intuition by bounding the difference of the gradient descent iterates $\vtheta_t$ and $\tilde \vtheta_t$ on the least squares loss $\mc L_{AB}(\vtheta)$ and the un-perturbed least-squares loss in~\eqref{eq:lossall}. 

\begin{theorem}
\label{thm:randproj}
Let $\mJ = \mU \mSigma \transp{\mV}$ be the singular value decomposition of $\mJ$, $\sigma_{\max}$ and $\sigma_{\min}$ are the largest and smallest singular values, and $\mU_{r}$ and $\mU_{n}$ are the left-singular vectors corresponding to the largest $r$ and the remaining singular values. 
Let $\vtheta_t$ and $\tilde \vtheta_t$ be the gradient descent iterates with stepsize $\eta$ starting at $\vtheta_0 =\tilde \vtheta_0= \mathbf{0}$ on the aforementioned least-squares problems.
\begin{enumerate}
\item[i)] Suppose that $t$ is sufficiently small so that \\$(1-\eta \sigma_{\min}^2)^t  \geq 1 - \frac{\sigma_{\min}^2}{\sigma_{r}^2}  \frac{\norm[2]{\transp{\mU}_{r}\vy}}{ \norm[2]{\transp{\mU}_{n}\vy} }$. 
Then, with probability at least $1- 4 t \exp\left(-\frac{\norm[F]{\mJ_A}^2}{2 \norm{\mJ_A}^2 } \right)$,
\begin{align*}
\norm[2]{\vtheta_{t} - \tilde \vtheta_{t}} 
\leq 8 \frac{\norm[F]{\mJ_A}}{\sqrt{s} \sigma_{r} } \frac{1}{ \sigma_{r} }
\norm[2]{\transp{\mU}_{r} \vy}.
\end{align*}
\item [ii)] 
Suppose that $\mJ$ has rank $r$. Then, with probability at least $1 - 2e^{-r^2}$, for all $t$, 
\begin{align*}
\norm[2]{\vtheta^{t} - \tilde \vtheta^{t}}
&\leq 
\frac{\sigma_{\max}}{\sigma_r} \sqrt{ c\frac{r}{s} }
 \frac{1}{\sigma_{r}}
\norm[2]{\tilde \vr_0}.
\end{align*}
\end{enumerate}
\end{theorem}

The theorem consists of two parts, both guaranteeing closeness of the iterates under slightly different assumptions.
 
The first part guarantees that if the sketch dimension is sufficiently large relative to the effective rank of the matrix $\mJ_A$, measured by $\norm[F]{\mJ_A}/\sigma_r$, then the solution obtained by applying gradient descent to the original and un-perturbed least-squares problems are very close. If the singular values decay quickly, then the statement ensures that after a certain number of iterations, sufficient to fitting the singular vectors corresponding to large singular values, the iterates are close.

The second part guarantees that the solution obtained by applying gradient descent to the original and un-perturbed least-squares problems are very close, provided that the sketch dimension is sufficiently large relative to $\sigma_{\max} \sqrt{r} / \sigma_r \geq \norm[F]{\mJ_A}/\sigma_r$. This is a slightly stronger requirement on the sketch dimension, but the closeness holds for all iterations $t$. 

There is a large body of literature which ensures that the solution of a randomly sketched least-squares problem behaves similarly to the solution obtained for the un-perturbed problem~\citep{sarlos_ImprovedApproximationAlgorithms_2006,agarwal_FastGlobalConvergence_2012,pilanci_RandomizedSketchesConvex_2015}. The proof of Theorem~\ref{thm:randproj} is conceptually similar to those of prior works, but differs as in our setup only part of the least-squares problem is sketched, and more importantly, our results also applies to over-parameterized models, unlike many previous results. 


\subsection{Continual learning on a sequence of two regression tasks}
\label{sec:contlearregression}

We next apply Theorem~\ref{thm:randproj} to obtain guarantees for continual learning of a regression task with the RSJ algorithm. We show that a sufficiently large sketch dimension provably enables continual learning. 

Suppose the data of task $T$, for $T \in \{A,B\}$ is generated by a Gaussian linear model 
\[
y = \innerprod{\vx}{\vtheta_T} + z,
\]
where $\vx \sim \mc N(0,\mI)$ and $z \sim \mc N(0,\sigma^2)$. 
Consider learning a linear model $f_\vtheta(\vx) = \innerprod{\vtheta}{\vx}$ with a quadratic loss. 
The risk if we draw a problem instance from one of the two tasks with equal probability is
\begin{align}
R(\vtheta) 
&= 
\frac{1}{2} \EX[(\vx,y) \sim P_A]{( \innerprod{\vtheta}{\vx} - y )^2} + \frac{1}{2} \EX[(\vx,y) \sim P_B]{( \innerprod{\vtheta}{\vx} - y )^2} \nonumber \\
&= \norm[2]{\vtheta - \vtheta_A}^2 + \norm[2]{\vtheta - \vtheta_B}^2 + \sigma^2. 
\label{eq:riskrregression}
\end{align}
Thus, the optimal linear model is $\vtheta^\ast = \frac{1}{2} (\vtheta_A + \vtheta_B)$. Suppose we obtain $n \gg d$ training examples from each of the two tasks and apply gradient descent until convergence on all of the data (i.e., the loss function in~\eqref{eq:lossall}). Then, after $t = O(\log(n) / \log(n/d))$ many gradient descent iterations (see supplement), the corresponding estimate $\tilde \vtheta_\iter$ obeys, with high probability,
\begin{align}
\label{eq:claim1eq}
\norm[2]{\tilde \vtheta_t - \vtheta^\ast}
\leq O\left( \sqrt{\frac{d}{n}} ( \norm[2]{\vtheta_A} +  \norm[2]{\vtheta_B} + \sigma)  \right).
\end{align}
If we learn both task sequentially with the RSJ algorithm, we get, by applying Theorem~\ref{thm:randproj}, that (see supplement for the details), with high probability, 
\begin{align}
\norm[2]{\vtheta_t - \vtheta^\ast}
&\leq O\left( \left( \sqrt{\frac{d}{n}} + \sqrt{ \frac{d}{s}} \right)  ( \norm[2]{\vtheta_A} +  \norm[2]{\vtheta_B} + \sigma)  \right). 
\label{eq:comperrors}
\end{align}
Thus, as long as we choose the dimension of the random projection on the order of $d$ (up to log-factors), the continual learning algorithm probably enables obtaining an estimator that has near-optimal risk, since this ensure that the RHS of~\eqref{eq:comperrors} is less than a small constant times $\norm[2]{\vtheta_A} +\norm[2]{\vtheta_B} + \sigma$, which applied to~\eqref{eq:riskrregression} implies a near-optimal risk (since 
$R(\vtheta_t) \leq 
R(\theta^\ast) 
+ 2 \norm[2]{\vtheta_t - \vtheta^\ast} 
\left(
\norm[2]{\vtheta_t - \vtheta^\ast}
+
\norm[2]{\vtheta_t - \vtheta_A}
+
\norm[2]{\vtheta_t - \vtheta_B}
\right)$). 

\subsection{Continual learning on more than two regression tasks}

The analysis from the previous section shows that if we learn two regression tasks sequentially, we obtain an estimate that is accurate up to two error terms (cf.~\eqref{eq:comperrors}): The first one is the statistical error due to learning on finite data. The second term is because we are not learning based on the original data of the first task $A$. 
If we move from two tasks to three and more the effect of those approximations compounds which makes it difficult to learn a large number of tasks sequentially. This explains why in the simulations shown previously for a linear random feature model (see~ Fig.~\ref{eq:permtaskrf}) 
 the gap between training  on all of the data and the RSJ algorithm increases in the number of tasks, and this gap becomes smaller if the approximation becomes better. 

\section{Guarantees for two-layer neural networks in the NTK regime}
\label{sec:NTK}
We now provide guarantees for wide neural networks in the so called neural-tangent-kernel (NTK) regime, in which the networks behave approximately linearly, as established by recent results 
~\citep{jacot_NeuralTangentKernel_2018,lee_DeepNeuralNetworks_2018a,du_GradientDescentProvably_2018b,oymak_ModerateOverparameterizationGlobal_2020}.

Consider a two-layer neural network with ReLU activation functions and $k$ neurons in the hidden layer:
\begin{align}
\label{eq:networkmodel}
f_{\vtheta}(\vx) 
= \frac{1}{\sqrt{k}} \relu(\transp{\vx} \mTheta) \vv.
\end{align}
Here, $\vx \in \reals^d$ is the input of the network, $\mTheta \in \reals^{d \times k}$ are the trainable weights of the first layer and $\vv \in \reals^k$ are fixed second-layer weights with the first half equal to $1$ and the second half equal to $-1$. The trainable parameters are the weight-matrix $\mTheta$, and we denote with $\vtheta$ the vectorized version of this matrix. 

The training data for task $T$, for $T \in \{A,B\}$, consists of $n$ points drawn iid from an (unknown) joint distribution $(\vx_i,y_i) \sim P_T$, and we assume for convenience that the data points are normalized ($\norm[2]{\vx}=1$) and that the labels are bounded ($|y_i| \leq 1$). 

Similarly as in the previous section, we consider the composite risk 
\begin{align}
\label{eq:riskntk}
R(\vtheta)
&=
\frac{1}{2} 
R_A(\vtheta)
+
\frac{1}{2}
R_B(\vtheta)
\end{align}
where
\begin{align*}
R_T(\vtheta) = \EX[(\vx,y) \sim P_T]{\loss(f_\vtheta(\vx) ,y )}.
\end{align*}
Here, $\loss \colon \reals \times \reals \to [0,1]$ is a loss function that is $1$-Lipschitz in its first argument and obeys $\loss(y,y) = 0$; an example of such a function is $\loss(z,y) = |z - y|$.
For the composite risk to be small, both the risks of task $A$ and $B$ have to be small. 

The following theorem quantifies the risk of the RSJ algorithm with the loss minimized via $\iter$-many iterations of gradient descent. 
Specifically, we first run gradient descent on the loss of task $A$, defined in equation~\eqref{eq:lossA} until convergence, and then run $t$ iterations on the RSJ-loss~\eqref{eq:lossAB}. 

Our risk bound depends on the Gram matrix $\mK \in \reals^{2n\times 2n}$ with entries defined as 
\[
[\mK]_{ij} = \frac{1}{2}
\left(
1 - \frac{\cos^{-1} \left( \innerprod{\vx_i}{\vx_j} \right)}{\pi}
\right)
\innerprod{\vx_i}{\vx_j},
\]
where $(i,j)$ are the pairs of training data points from the two tasks.

\begin{theorem}
\label{thm:ntk}
Let $\alpha>0$ be the smallest singular value of the Gram matrix $\mK$, and consider the network in the NTK regime where $k\to \infty$. 
Let $f_{\vtheta_\iter}$ be the network trained with $\iter$ iterations of gradient descent applied to the loss of the RSJ algorithm. 
Then with probability at least 
$1- 4 t \exp( - \frac{\norm[F]{\mJ_A}^2}{2 \norm{\mJ_A}^2 } )$, the risk of the network trained with at least $\iter \geq \frac{\log(1 - \eta \alpha)}{ \log(1/n) }$ gradient iterations is bounded by 
\begin{align}
\risk(\vtheta_\iter)
&\leq
2\sqrt{
\frac{1}{n}
\transp{\vy} \inv{\mK} \vy
}  +
\frac{3}{\sqrt{n}}  
+
\frac{1}{\sqrt{s} \alpha^2}
\left(
10 \norm[F]{\mJ_A}
+
\frac{\norm[F]{ \mK} }{ \sqrt{n} } 
\right),
\label{eq:ntkriskbound}
\end{align}
where $\vy = [\vy_A, \vy_B] \in \reals^{2n}$ contains the labels of the training data of the two tasks. 
\end{theorem}
The theorem establishes that the risk is bounded by a complexity measure of the data defined as $\sqrt{ \frac{1}{n} \transp{\vy} \inv{\mK} \vy }$, plus two perturbation terms, one of which depends on the quality of the random projection.
The complexity measure of the data has been studied by~\citet{arora_FineGrainedAnalysisOptimization_2019}, and measures whether the training set of tasks $A$ and $B$ can be well represented with the kernel associated with the relu-network. If those training sets can be well represented with this kernel, the complexity measure $\sqrt{
\frac{1}{n}
\transp{\vy} \inv{\mK} \vy
}$ is small. 
See \citep{arora_FineGrainedAnalysisOptimization_2019} for a detailed discussion on the interpretation of this complexity measure. 

For our purpose, it is important to note that if we were to train the neural network on the data from task $A$ and $B$ simultaneously, we would obtain the same risk bound as above but without the perturbation term in equation~\eqref{eq:ntkriskbound} starting with $\frac{1}{\sqrt{s}\alpha^2}$. The perturbation term is small if the dimension of the random projection is on the order of the effective rank of the Jacobian $\mJ_A$; this term can be computed from the data and thus we can verify whether we are choosing the random projection dimension $s$ sufficiently large.


\section{Negative results for learning Gaussian mixture models sequentially with EWC}
\label{sec:EWCtheory}

In this section we study the popular data permutation experiment theoretically, and show why EWC and L2-regularization \emph{can} perform well even in a setup where the diagonal of the Jacobian outer product is a poor approximation of the original Jacobian outer product (recall the experiment in Figure~\ref{eq:permtask}). 

Consider a binary version of the MNIST permutation experiment. Task A is to classify the digits $\{0,1\}$ from the original feature vectors (i.e., the vectorized $28 \times 28$ pixel images), task $B$ to classify the digits $\{0,1\}$ from the vectorized images permuted with a random perturbation, fixed for task $B$, and task $C$ is generated in the same way, with a new permutation. 

A simple mathematical abstraction for this is a Gaussian mixture model where the class means of the two classes point in different directions. 
In the MNIST permutation experiment, task B is obtained by simply shuffling each of the feature vectors (the pixels) with a random permutation. 

\begin{figure}[t]
\begin{center}
\begin{tikzpicture}[>=latex,scale=1.2]

\draw[->] (-2,0) -- (2,0);
\draw[->] (0,-2) -- (0,2);   

\node at (0,1.3) {\includegraphics[width=0.8cm]{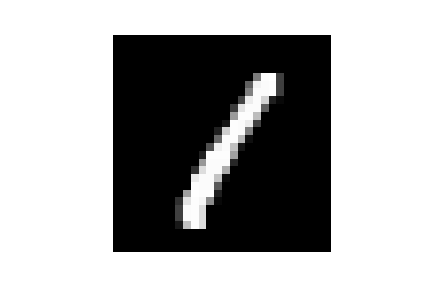}};
\node at (0,1.6) {\includegraphics[width=0.8cm]{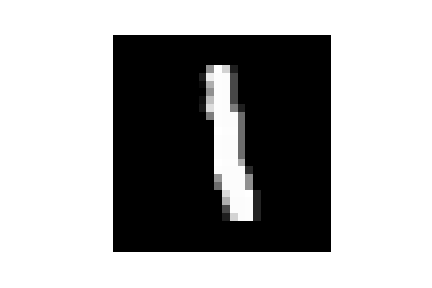}};
\node at (0,1.0) {\includegraphics[width=0.8cm]{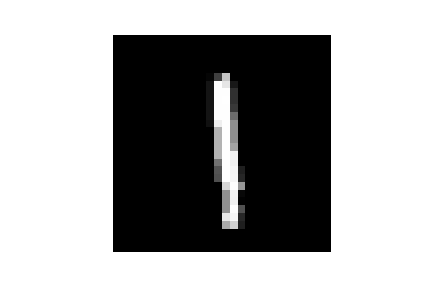}};
\node at (0.3,1.3) {\includegraphics[width=0.8cm]{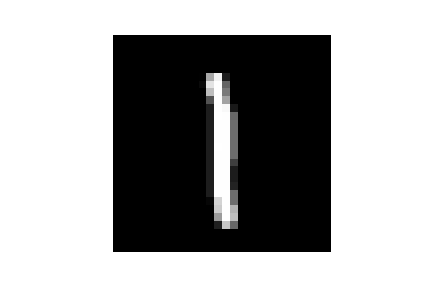}};
\node at (-0.3,1.3) {\includegraphics[width=0.8cm]{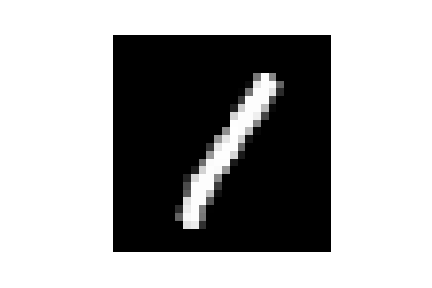}};

\node at (0,-1.3) {\includegraphics[width=0.8cm]{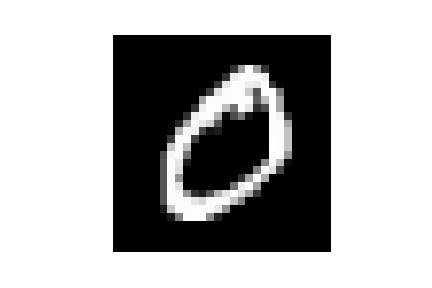}};
\node at (0,-1.6) {\includegraphics[width=0.8cm]{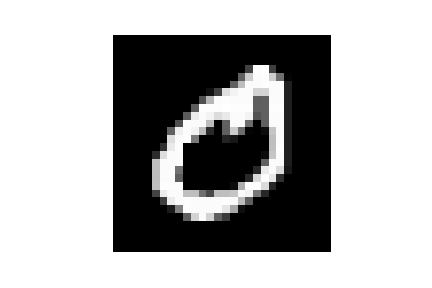}};
\node at (0,-1.0) {\includegraphics[width=0.8cm]{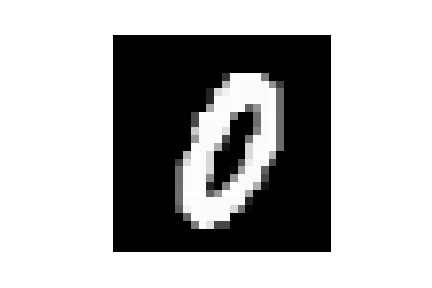}};
\node at (0.3,-1.3) {\includegraphics[width=0.8cm]{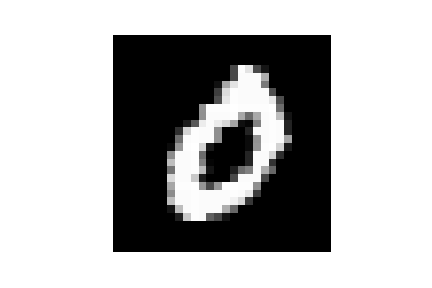}};
\node at (-0.3,-1.3) {\includegraphics[width=0.8cm]{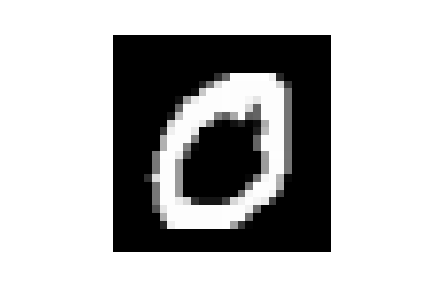}};

\node at (1.3,0) {\includegraphics[width=0.8cm]{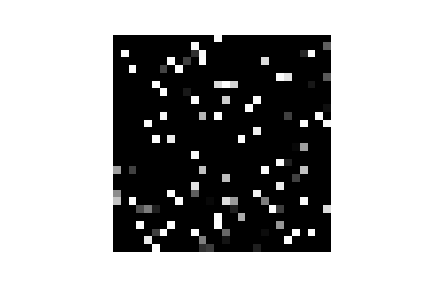}};
\node at (1.6,0) {\includegraphics[width=0.8cm]{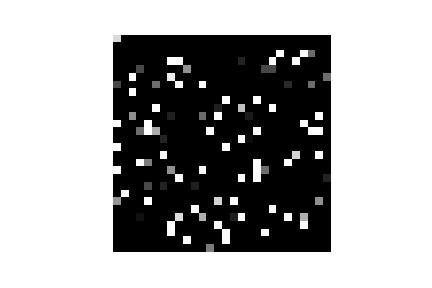}};
\node at (1.0,0) {\includegraphics[width=0.8cm]{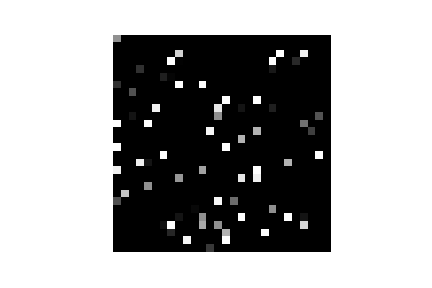}};
\node at (1.3,0.3) {\includegraphics[width=0.8cm]{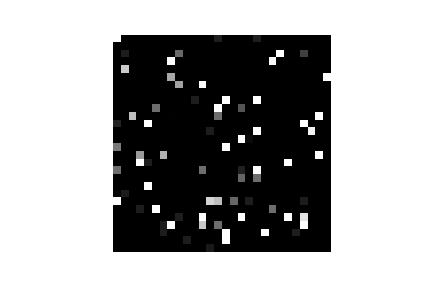}};
\node at (1.3,-0.3) {\includegraphics[width=0.8cm]{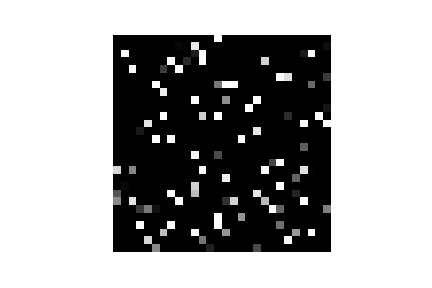}};

\node at (-1.3,0) {\includegraphics[width=0.8cm]{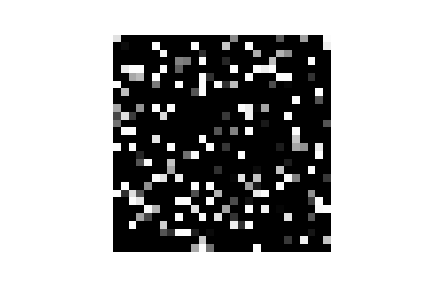}};
\node at (-1.6,0) {\includegraphics[width=0.8cm]{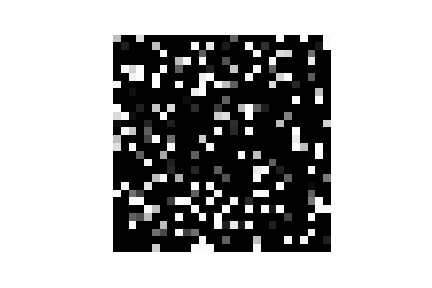}};
\node at (-1.0,0) {\includegraphics[width=0.8cm]{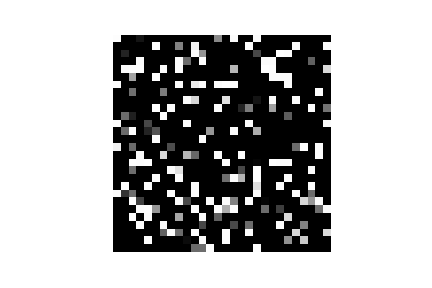}};
\node at (-1.3,0.3) {\includegraphics[width=0.8cm]{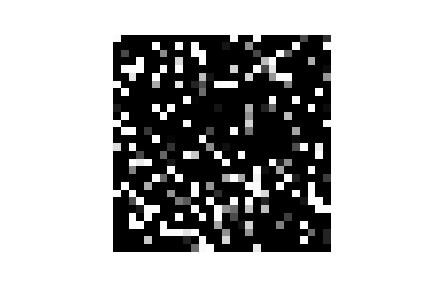}};
\node at (-1.3,-0.3) {\includegraphics[width=0.8cm]{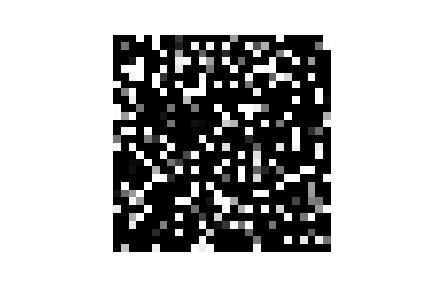}};

\end{tikzpicture}
\end{center}

\vspace{-0.6cm}

\caption{
\label{example:permutationtask}
A visualization of the permutation task: The first axis shows task A of distinguishing ones and zeros. The second axis shows task B of distinguishing shuffled ones and zeros (each image is shuffled with the same random permutation). Due to the random shuffling, the two axis are approximately orthogonal. We therefore model this with a Gaussian mixture model where the two tasks are orthogonal to each other.
}
\end{figure}

This can be modeled with a new task $B$ of the Gaussian mixture model with class means that are near-orthogonal to the means of task $A$. 
This is illustrated in Figure~\ref{example:permutationtask}. 
Motivated by this observation, we study the following Gaussian mixture model for continual learning.

 
\paragraph{Gaussian mixture model for continual learning:}
Consider the standard binary classification setting, where the data for each task is distributed as a mixture of two Gaussians. 
For task $T=A, B, \ldots$, the response $y \in \{-1,1\}$ is uniformly distributed, and the feature vector $\vx$ given the class label $y$ is distributed as 
$
\vx| y \sim \mc N( y \vmu_T, \sigma^2\mI ).
$
Here, $\vmu_T \in \reals^{d}$ is a fixed class mean vector with unit norm, and $\sigma^2 > 0$ is the within-class variance. The Bayes optimal classifier for a given task is $\hat y(\vx) = \mathrm{sign}(\innerprod{\vmu_T}{\vx})$. 

Consider the linear classifier $\hat y_\vtheta(\vx) = \mathrm{sign}(\innerprod{\vtheta}{\vx})$. The corresponding Bayes risk on task $T$ is
\begin{align*}
R_T(\vtheta) 
=
\PR[(\vx,y) \sim P_T]{\ind{ \hat y(\vx) \neq y }} 
=
\Phi\left( 
\frac{ \innerprod{\vtheta}{\vmu_T} }{\norm[2]{\vtheta} \sigma}
\right),
\end{align*}
where $\Phi(x) = (2\pi)^{-1/2}\int_{-\infty}^x e^{-t^2/2} dt$ is the cumulative distribution function of the standard normal distribution. 
Our goal is to learn the tasks sequentially so that the risk of a task chosen with equal probability is small. Specifically, we wish the risk 
\begin{align*}
R_{A:T}(\vtheta) = \frac{1}{T} 
\left(
R_A(\vtheta) + R_B(\vtheta) + \ldots + R_T(\vtheta)
\right)
\end{align*}
to be small. For simplicity, we consider two tasks $A$ and $B$. The optimal $\vtheta$ that minimizes the risk of tasks $A$ and $B$, i.e., $R_{AB}$ is 
is a linear combination of the two optimal points for the individual tasks $\vtheta_{AB}^\ast = \vmu_A + \vmu_B$. 

We next identify a setup in which the EWC algorithm and L2 regularization provably succeed in learning tasks $A$ and $B$ sequentially. The EWC algorithm in our formulation is trained with the quadratic loss, and not with another suitable loss such as cross-entropy, and we consider the infinite-data case, i.e., we study EWC and L2 applied to the population risk, so that we do not have to take finite-data effects into account.

\begin{theorem}
\label{thm:cansucceed}
Suppose that the class means $\vmu_A,\vmu_B \in \reals^d$ lie on a hypercube, i.e., each entry has equal magnitude $1/d$ and that the inner product between the class means is non-negative ($\innerprod{\vmu_A}{\vmu_B} \geq 0$). Then: 
\begin{enumerate}
\item[i)] There is a choice of regularization parameter $\lambda$, such that EWC is optimal (i.e., its solution minimizes the Bayes risk).
\item[ii)] There is a choice of regularization parameter $\lambda$, such that L2 is optimal.
\end{enumerate}
\end{theorem}

This result allows the diagonal of the Jacobian outer product (i.e., the Hessian for this setup) to be a poor approximation of the Jacobian outer product. 
Thus, perhaps surprisingly, there are setups where EWC (and L2-regularization) provably succeeds even when the penalty of EWC (or $L2$-regularization) is a poor approximation of the loss on past data. 

The MNIST permutation problem approximately corresponds to a setup where the class mean are on a hypercube (because most pixels are one or zero) and therefore the theorem gives a potential justification why EWC works so well on this problem. 

At the same time, there exist, as expected, problem instances where EWC and L2 regularization provably fail at continual learning, even if given infinitely many training examples:

\begin{theorem}
\label{thm:canfail}
There are problem instance of the Gaussian mixture model in $\reals^d, d\geq3$ (i.e., a choice of $\vmu_A,\vmu_B$, and $\sigma^2$) such that:
\begin{enumerate}
\item[i)] The risk of EWC for all values of the regularization parameter $\lambda$ is at least $3/2$-times the optimal risk: $R_{AB}(\vtheta_{EWC}) \geq 3/2 R_{AB}(\vtheta_{AB}^\ast)$. 
\item[ii)] The risk of L2-regularization for all values of the regularization parameter $\lambda$ is $3/2$-times the optimal risk: $R_{AB}(\vtheta_{L2}) \geq 3/2 R_{AB}(\vtheta_{AB}^\ast)$. 
\end{enumerate}
\end{theorem}

\section{Conclusion}
In this paper we studied a family of algorithms in order to understand regularization based continual learning algorithms. We showed that popular regularization-based learning algorithms, in particular EWC, can provably fail if the regularization does not approximate the loss for past data well. We also showed that this can be fixed by working with better approximations of the Jacobian. However, while the resulting algorithm provably succeeds, it might comes at a high memory cost. Our results indicate that regularization-based continual learning algorithms that provably succeed for a variety of setups might need to have a large memory footprint in return. 

\section*{Acknowledgements}

Many thanks to Paul Hand for discussions about continual learning, and to Daniel LeJeune for proofreading the manuscript. 
RH is supported by the Institute of Advanced Studies at the Technical University of Munich, and the Deutsche Forschungsgemeinschaft (DFG, German Research Foundation) - 456465471, 464123524, and by the NSF under award IIS-1816986. 

\section*{Code}

Code to reproduce the experiments in this paper is at \\ \href{https://github.com/MLI-lab/regularization_based_continual_learning}{https://github.com/MLI-lab/regularization\_based\_continual\_learning}.


\printbibliography

@article{agarwal_FastGlobalConvergence_2012,
  title = {Fast Global Convergence of Gradient Methods for High-Dimensional Statistical Recovery},
  author = {Agarwal, Alekh and Negahban, Sahand and Wainwright, Martin J.},
  year = {2012},
  journal = {Annals of Statistics},
  volume = {40},
  number = {5},
  pages = {2452--2482},
  publisher = {{Institute of Mathematical Statistics}},
  langid = {english},
  mrnumber = {MR3097609},
  zmnumber = {1373.62244}
}

@inproceedings{alquier_RegretBoundsLifelong_2017,
  title = {Regret {{Bounds}} for {{Lifelong Learning}}},
  booktitle = {International {{Conference}} on {{Artificial Intelligence}} and {{Statistics}}},
  author = {Alquier, Pierre and Mai, The Tien and Pontil, Massimiliano},
  year = {2017},
  pages = {261--269},
  langid = {english}
}

@inproceedings{arora_FineGrainedAnalysisOptimization_2019,
  title = {Fine-{{Grained Analysis}} of {{Optimization}} and {{Generalization}} for {{Overparameterized Two}}-{{Layer Neural Networks}}},
  booktitle = {International {{Conference}} on {{Machine Learning}}},
  author = {Arora, Sanjeev and Du, Simon and Hu, Wei and Li, Zhiyuan and Wang, Ruosong},
  year = {2019},
  pages = {322--332}
}

@article{bennani_GeneralisationGuaranteesContinual_2020b,
  title = {Generalisation {{Guarantees}} for {{Continual Learning}} with {{Orthogonal Gradient Descent}}},
  author = {Bennani, Mehdi Abbana and Doan, Thang and Sugiyama, Masashi},
  year = {2020},
  journal = {arXiv:2006.11942 [cs, stat]}
}

@inproceedings{chaudhry_RiemannianWalkIncremental_2018,
  title = {Riemannian {{Walk}} for {{Incremental Learning}}: Understanding {{Forgetting}} and {{Intransigence}}},
  booktitle = {Proceedings of the {{European Conference}} on {{Computer Vision}} ({{ECCV}})},
  author = {Chaudhry, Arslan and Dokania, Puneet K. and Ajanthan, Thalaiyasingam and Torr, Philip H. S.},
  year = {2018},
  pages = {532--547}
}

@inproceedings{doan_TheoreticalAnalysisCatastrophic_2021,
  title = {A {{Theoretical Analysis}} of {{Catastrophic Forgetting}} through the {{NTK Overlap Matrix}}},
  booktitle = {International {{Conference}} on {{Artificial Intelligence}} and {{Statistics}}},
  author = {Doan, Thang and Bennani, Mehdi Abbana and Mazoure, Bogdan and Rabusseau, Guillaume and Alquier, Pierre},
  year = {2021},
  pages = {1072--1080},
  langid = {english}
}

@inproceedings{du_GradientDescentProvably_2018b,
  title = {Gradient {{Descent Provably Optimizes Over}}-Parameterized {{Neural Networks}}},
  booktitle = {International {{Conference}} on {{Learning Representations}}},
  author = {Du, Simon S. and Zhai, Xiyu and Poczos, Barnabas and Singh, Aarti},
  year = {2018}
}

@article{french_CatastrophicForgettingConnectionist_1999,
  title = {Catastrophic Forgetting in Connectionist Networks},
  author = {French, Robert M.},
  year = {1999},
  journal = {Trends in Cognitive Sciences},
  volume = {3},
  number = {4},
  pages = {128--135},
  langid = {english}
}

@article{goodfellow_EmpiricalInvestigationCatastrophic_2015,
  title = {An {{Empirical Investigation}} of {{Catastrophic Forgetting}} in {{Gradient}}-{{Based Neural Networks}}},
  author = {Goodfellow, Ian J. and Mirza, Mehdi and Xiao, Da and Courville, Aaron and Bengio, Yoshua},
  year = {2015},
  journal = {arXiv:1312.6211 [cs, stat]}
}

@inproceedings{heckel_EarlyStoppingDeep_2021,
  title = {Early {{Stopping}} in {{Deep Networks}}: Double {{Descent}} and {{How}} to {{Eliminate}} It},
  booktitle = {International {{Conference}} on {{Learning Representations}}},
  author = {Heckel, Reinhard and Yilmaz, Fatih Furkan},
  year = {2021}
}

@article{hsu_ReevaluatingContinualLearning_2019,
  title = {Re-Evaluating {{Continual Learning Scenarios}}: A {{Categorization}} and {{Case}} for {{Strong Baselines}}},
  author = {Hsu, Yen-Chang and Liu, Yen-Cheng and Ramasamy, Anita and Kira, Zsolt},
  year = {2019},
  journal = {arXiv:1810.12488 [cs]}
}

@inproceedings{jacot_NeuralTangentKernel_2018,
  title = {Neural {{Tangent Kernel}}: Convergence and {{Generalization}} in {{Neural Networks}}},
  booktitle = {Neural {{Information Processing Systems}}},
  author = {Jacot, Arthur and Gabriel, Franck and Hongler, Clement},
  year = {2018},
  pages = {8571--8580}
}

@inproceedings{kemker_MeasuringCatastrophicForgetting_2018,
  title = {Measuring {{Catastrophic Forgetting}} in {{Neural Networks}}},
  booktitle = {Proceedings of the {{AAAI Conference}} on {{Artificial Intelligence}}},
  author = {Kemker, Ronald and McClure, Marc and Abitino, Angelina and Hayes, Tyler and Kanan, Christopher},
  year = {2018},
  volume = {32},
  langid = {english}
}

@article{kirkpatrick_OvercomingCatastrophicForgetting_2017,
  title = {Overcoming Catastrophic Forgetting in Neural Networks},
  author = {Kirkpatrick, James and Pascanu, Razvan and Rabinowitz, Neil and Veness, Joel and Desjardins, Guillaume and Rusu, Andrei A. and Milan, Kieran and Quan, John and Ramalho, Tiago and {Grabska-Barwinska}, Agnieszka and Hassabis, Demis and Clopath, Claudia and Kumaran, Dharshan and Hadsell, Raia},
  year = {2017},
  journal = {Proceedings of the National Academy of Sciences},
  volume = {114},
  number = {13},
  pages = {3521--3526},
  langid = {english}
}

@inproceedings{kunstner_LimitationsEmpiricalFisher_2019,
  title = {Limitations of the {{Empirical Fisher Approximation}} for {{Natural Gradient Descent}}},
  booktitle = {Neural {{Information Processing Systems}}},
  author = {Kunstner, Frederik and Balles, Lukas and Hennig, Philipp},
  year = {2019}
}

@inproceedings{lee_DeepNeuralNetworks_2018a,
  title = {Deep {{Neural Networks}} as {{Gaussian Processes}}},
  booktitle = {International {{Conference}} on {{Learning Representations}}},
  author = {Lee, Jaehoon and Bahri, Yasaman and Novak, Roman and Schoenholz, Samuel S. and Pennington, Jeffrey and {Sohl-Dickstein}, Jascha},
  year = {2018}
}

@article{li_LearningForgetting_2018,
  title = {Learning without {{Forgetting}}},
  author = {Li, Z. and Hoiem, D.},
  year = {2018},
  journal = {IEEE Transactions on Pattern Analysis and Machine Intelligence},
  volume = {40},
  number = {12},
  pages = {2935--2947}
}

@inproceedings{li_LifelongLearningSketched_2021,
  title = {Lifelong {{Learning}} with {{Sketched Structural Regularization}}},
  booktitle = {Asian {{Conference}} on {{Machine Learning}}},
  author = {Li, Haoran and Krishnan, Aditya and Wu, Jingfeng and Kolouri, Soheil and Pilly, Praveen K. and Braverman, Vladimir},
  year = {2021}
}

@inproceedings{liu_RotateYourNetworks_2018,
  title = {Rotate Your {{Networks}}: Better {{Weight Consolidation}} and {{Less Catastrophic Forgetting}}},
  booktitle = {2018 24th {{International Conference}} on {{Pattern Recognition}} ({{ICPR}})},
  author = {Liu, X. and Masana, M. and Herranz, L. and de Weijer, J. Van and L{\'o}pez, A. M. and Bagdanov, A. D.},
  year = {2018},
  pages = {2262--2268}
}

@book{mohri_FoundationsMachineLearning_2012,
  title = {Foundations of {{Machine Learning}}},
  author = {Mohri, Mehryar and Rostamizadeh, Afshin and Talwalkar, Ameet},
  year = {2012},
  publisher = {{MIT Press}},
  isbn = {978-0-262-30473-3},
  langid = {english}
}

@article{oymak_ModerateOverparameterizationGlobal_2020,
  title = {Towards Moderate Overparameterization: Global Convergence Guarantees for Training Shallow Neural Networks},
  author = {Oymak, Samet and Soltanolkotabi, Mahdi},
  year = {2020},
  journal = {IEEE Journal on Selected Areas in Information Theory},
  langid = {english}
}

@article{pan_ContinualDeepLearning_2021,
  title = {Continual {{Deep Learning}} by {{Functional Regularisation}} of {{Memorable Past}}},
  author = {Pan, Pingbo and Swaroop, Siddharth and Immer, Alexander and Eschenhagen, Runa and Turner, Richard E. and Khan, Mohammad Emtiyaz},
  year = {2021},
  journal = {arXiv:2004.14070 [cs, stat]}
}

@article{pilanci_RandomizedSketchesConvex_2015,
  title = {Randomized {{Sketches}} of {{Convex Programs With Sharp Guarantees}}},
  author = {Pilanci, M. and Wainwright, M. J.},
  year = {2015},
  journal = {IEEE Transactions on Information Theory},
  volume = {61},
  number = {9},
  pages = {5096--5115}
}

@inproceedings{rebuffi_ICaRLIncrementalClassifier_2017,
  title = {{{iCaRL}}: Incremental {{Classifier}} and {{Representation Learning}}},
  booktitle = {Proceedings of the {{IEEE Conference}} on {{Computer Vision}} and {{Pattern Recognition}}},
  author = {Rebuffi, Sylvestre-Alvise and Kolesnikov, Alexander and Sperl, Georg and Lampert, Christoph H.},
  year = {2017},
  pages = {2001--2010}
}

@article{ritter_OnlineStructuredLaplace_2018,
  title = {Online {{Structured Laplace Approximations}} for {{Overcoming Catastrophic Forgetting}}},
  author = {Ritter, Hippolyt and Botev, Aleksandar and Barber, David},
  year = {2018},
  journal = {Advances in Neural Information Processing Systems},
  volume = {31},
  pages = {3738--3748},
  langid = {english}
}

@article{robins_CatastrophicForgettingRehearsal_1995,
  title = {Catastrophic {{Forgetting}}, {{Rehearsal}} and {{Pseudorehearsal}}},
  author = {Robins, Anthony},
  year = {1995},
  journal = {Connection Science},
  volume = {7},
  number = {2},
  pages = {123--146},
  publisher = {{Taylor \& Francis}}
}

@inproceedings{rudelson_NonasymptoticTheoryRandom_2010,
  title = {Non-Asymptotic Theory of Random Matrices: Extreme Singular Values},
  booktitle = {Proceedings of the {{International Congress}} of {{Mathematicians}}},
  author = {Rudelson, Mark and Vershynin, Roman},
  year = {2010},
  volume = {3},
  pages = {1576--1602}
}

@article{rusu_ProgressiveNeuralNetworks_2016,
  title = {Progressive {{Neural Networks}}},
  author = {Rusu, Andrei A. and Rabinowitz, Neil C. and Desjardins, Guillaume and Soyer, Hubert and Kirkpatrick, James and Kavukcuoglu, Koray and Pascanu, Razvan and Hadsell, Raia},
  year = {2016},
  journal = {arXiv:1606.04671 [cs]}
}

@inproceedings{ruvolo_ELLAEfficientLifelong_2013,
  title = {{{ELLA}}: An {{Efficient Lifelong Learning Algorithm}}},
  booktitle = {International {{Conference}} on {{Machine Learning}}},
  author = {Ruvolo, Paul and Eaton, Eric},
  year = {2013},
  pages = {507--515},
  langid = {english}
}

@inproceedings{sarlos_ImprovedApproximationAlgorithms_2006,
  title = {Improved {{Approximation Algorithms}} for {{Large Matrices}} via {{Random Projections}}},
  booktitle = {{{IEEE Symposium}} on {{Foundations}} of {{Computer Science}} ({{FOCS}}'06)},
  author = {Sarlos, Tamas},
  year = {2006},
  pages = {143--152}
}

@inproceedings{schwarz_ProgressCompressScalable_2018,
  title = {Progress \& {{Compress}}: A Scalable Framework for Continual Learning},
  booktitle = {International {{Conference}} on {{Machine Learning}}},
  author = {Schwarz, Jonathan and Czarnecki, Wojciech and Luketina, Jelena and {Grabska-Barwinska}, Agnieszka and Teh, Yee Whye and Pascanu, Razvan and Hadsell, Raia},
  year = {2018},
  pages = {4528--4537},
  langid = {english}
}

@inproceedings{shin_ContinualLearningDeep_2017,
  title = {Continual {{Learning}} with {{Deep Generative Replay}}},
  booktitle = {Neural {{Information Processing Systems}}},
  author = {Shin, Hanul and Lee, Jung Kwon and Kim, Jaehong and Kim, Jiwon},
  year = {2017},
  volume = {30}
}

@article{vandeven_ThreeScenariosContinual_2019,
  title = {Three Scenarios for Continual Learning},
  author = {{van de Ven}, Gido M. and Tolias, Andreas S.},
  year = {2019},
  journal = {arXiv:1904.07734 [cs, stat]}
}

@article{yin_OptimizationGeneralizationRegularizationBased_2020,
  title = {Optimization and {{Generalization}} of {{Regularization}}-{{Based Continual Learning}}: A {{Loss Approximation Viewpoint}}},
  author = {Yin, Dong and Farajtabar, Mehrdad and Li, Ang and Levine, Nir and Mott, Alex},
  year = {2020},
  journal = {arXiv:2006.10974 [cs, stat]}
}

@inproceedings{zenke_ContinualLearningSynaptic_2017,
  title = {Continual {{Learning Through Synaptic Intelligence}}},
  booktitle = {International {{Conference}} on {{Machine Learning}}},
  author = {Zenke, Friedemann and Poole, Ben and Ganguli, Surya},
  year = {2017},
  pages = {3987--3995},
  langid = {english}
}

\newpage

\appendix

\onecolumn
\section{Proof of Proposition~\ref{eq:proplin}}

For a linear model, the Jacobian $\mJ_T \in \reals^{n\times \pdim}$ contains the feature maps $\psi(\vx_i)$ as rows. 
With this notation, the loss function of task $A$ becomes
\begin{align*}
\mc L_A(\vtheta)
=
\frac{1}{2} \norm[2]{\mJ_A \vtheta - \vy_A}^2,
\end{align*}
where $\vy_A \in \reals^n$ are the responses of task $A$. The minimizer $\vtheta_A$ obeys
\begin{align*}
\transp{\mJ}_A \mJ_A \vtheta_A = \transp{\mJ}_A \vy_A. 
\end{align*}
Next, consider learning $B$ after $A$ and note that
\begin{align*}
\min_{\vtheta} \mc L_{AB}(\vtheta)
&=
\min_{\vtheta}
\frac{1}{2}
\transp{(\mJ_B \vtheta - \vy_B)}
(\mJ_B \vtheta - \vy_B) 
+
\frac{\lambda}{2} \transp{(\vtheta - \vtheta_A)} 
\transp{\mJ}_A\mJ_A 
(\vtheta - \vtheta_A). \\
&=
\min_{\vtheta}
\frac{1}{2}
\norm[2]{\mJ_B \vtheta - \vy_B }^2
+ 
\frac{\lambda}{2} \transp{\vtheta}
\transp{\mJ}_A\mJ_A \vtheta
-
\lambda \transp{\vtheta}
\transp{\mJ}_A\mJ_A
\transp{\vtheta}_A \\
&=
\min_{\vtheta}
\frac{1}{2}
\norm[2]{\mJ_B \vtheta - \vy_B }^2
+ 
\frac{\lambda}{2} \transp{\vtheta}
\transp{\mJ}_A\mJ_A \vtheta
-
\lambda \transp{\vtheta}
\transp{\mJ}_A\vy_A \\
&=
\min_{\vtheta}
\frac{1}{2}
\norm[2]{\mJ_B \vtheta - \vy_B }^2
+ 
\frac{\lambda}{2} \norm[2]{\mJ_A \vtheta - \vy_A}^2.
\end{align*}
The analogous argument shows that the minimizer of $\mc L_{ABC}(\vtheta)$ is equal to the minimizer of $\mc L_A(\vtheta) + \mc L_B(\vtheta)+\mc L_C(\vtheta)$, and likewise for more tasks.


\section{Proof of Theorem~\ref{thm:randproj}: Analysis for sketched least-squares}

In this section we prove Theorem~\ref{thm:randproj} by showing that under certain conditions, the solution to a sketched least squares is close to that of an associated non-sketched one.

\subsection{Proof of Theorem~\ref{thm:randproj}, part i}

We consider the least-squares objective
\begin{align*}
\mc L_{AB}(\vtheta) 
&=
\mc L_{B}(\vtheta) + \frac{1}{2} \transp{(\vtheta - \vtheta_A)} \transp{\mJ}_A \transp{\mS} \mS \mJ_A(\vtheta - \vtheta_A) \\
&=
\frac{1}{2}
\transp{\vtheta} \transp{\mJ}_B \mJ_B \vtheta 
- \transp{\vtheta} \transp{\mJ}_B \vy_B
+
\frac{1}{2}
\transp{\vtheta} \transp{\mJ}_A \transp{\mS} \mS \mJ_A \vtheta 
-
\transp{\vtheta} \transp{\mJ}_A \transp{\mS} \mS \mJ_A\vtheta_A
 + c \\
&=
\frac{1}{2}
\transp{\vtheta} 
\begin{bmatrix}
\transp{\mJ}_A\transp{\mS}, \transp{\mJ_B}
\end{bmatrix}
\begin{bmatrix}
\mS \mJ_A \\ \mJ_B
\end{bmatrix}
\vtheta
-
\transp{\vtheta} 
\begin{bmatrix}
\transp{\mJ}_A\transp{\mS}, \transp{\mJ_B}
\end{bmatrix}
\begin{bmatrix}
\mS \mJ_A \vtheta_A \\ \vy_B
\end{bmatrix}
 + c \\
&=
\frac{1}{2}
\transp{\vtheta} \transp{\mJ} \mP
\mJ
\vtheta
-
\transp{\vtheta} 
\transp{\mJ}
\mP \vy
 + c.
\end{align*}
Here, $c$ is a numerical constant, independent of the optimization parameter $\vtheta$, and we defined
\begin{align}
\label{eq:defJ}
\mJ 
= 
\begin{bmatrix}
\mJ_A \\ \mJ_B
\end{bmatrix},
\quad
\mP = \begin{bmatrix}
\transp{\mS} \mS & \mathbf{0} \\ \mathbf{0} & \mI
\end{bmatrix},
\quad
\vy
=
\begin{bmatrix}
\mJ_A \vtheta_A \\ \vy_B
\end{bmatrix},
\end{align}
for notational convenience. 

The gradient descent iterates with stepsize $\eta$ for minimizing the loss $\mc L_{AB}(\vtheta)$ are
\begin{align*}
\vtheta^{t+1} = \vtheta^t - \eta ( \transp{\mJ} \mP \mJ \vtheta^t - \transp{\mJ} \mP \vy).
\end{align*}
We bound the difference to the gradient iterates for minimizing the least squares objective without the random projection matrix
\begin{align*}
\tilde{\mc L}_{AB}(\vtheta) 
&=
\frac{1}{2}
\transp{\vtheta} \transp{\mJ} \mJ
\vtheta
+
\transp{\vtheta} 
\transp{\mJ} \vy.
\end{align*}
The gradients descent iterates for minimizing the loss $\tilde{\mc L}_{AB}(\vtheta)$ are 
\begin{align*}
\tilde \vtheta^{t+1} = \tilde \vtheta^t - \eta ( \transp{\mJ} \mJ \tilde \vtheta^t - \transp{\mJ} \vy).
\end{align*}
The following lemma bounds the deviation of the two versions of gradient descent. 
\begin{lemma}
\label{lem:lemdiffiterates}
Let $\mJ = \mU \mSigma \transp{\mV}$ be the singular value decomposition of $\mJ$, let $\sigma_{\max}$ and $\sigma_{\min}$ the largest and smallest singular values, and let $\mU_{r}$ and $\mU_{n}$ be the left-singular vectors corresponding to the $r$-largest and the remaining singular values. 
Let $\vtheta_t$ and $\tilde \vtheta_t$ be the gradient descent iterates after $t$ iterations starting at $\vtheta_0 = \tilde \vtheta_0= \mathbf{0}$. 
With probability at least $1- 4 t \exp( - \frac{\norm[F]{\mJ_A}^2}{2 \norm{\mJ_A}^2 } )$ over the random sketch $\mS \in \reals^{s\times n}$, the difference of the iterates of the original and the sketched problem is bounded by
\begin{align*}
\norm[2]{\vtheta^{t} - \tilde \vtheta^{t}} 
&\leq 
5 \frac{\norm[F]{\mJ_A}}{\sqrt{s}}
\left(
 \frac{1 - (1-\eta \sigma_{r}^2)^t}{ \sigma_{r}^2}
\norm[2]{\transp{\mU}_{r} \vy}
+
\frac{1 - (1-\eta \sigma_{\min}^2)^t}{ \sigma_{\min}^2}
\norm[2]{ \transp{\mU}_{n} \vy}
\right).
\end{align*}
\end{lemma}
Theorem~\ref{thm:randproj} follows from the lemma, by using the assumption $(1-\eta \sigma_{\min}^2)^t  \geq 1 - \frac{\sigma_{\min}^2}{\sigma_{r}^2}  \frac{\norm[2]{\transp{\mU}_{r}\vy}}{ \norm[2]{\transp{\mU}_{n}\vy} }$ to conclude 
\begin{align*}
\norm[2]{\vtheta^{t} - \tilde \vtheta^{t}} 
&\leq
5 \frac{\norm[F]{\mJ}}{\sqrt{s}}
\left(
\frac{1}{ \sigma_{r}^2}
\norm[2]{\transp{\mU}_{r} \vr_0}
+
\frac{1}{ \sigma_{\min}^2}
\norm[2]{ \transp{\mU}_{n} \vr_0} 
\frac{\sigma_{\min}^2}{\sigma_{r}^2}  \frac{\norm[2]{\transp{\mU}_{r}\vy}}{ \norm[2]{\transp{\mU}_{n}\vy} }
\right) \\
&=
5 \frac{\norm[F]{\mJ}}{\sqrt{s}} \frac{1}{ \sigma_{r}^2}
\norm[2]{\transp{\mU}_{r} \vr_0}.
\end{align*}
In the reminder of this section we prove Lemma~\ref{lem:lemdiffiterates}.


\subsection{Proof of Lemma~\ref{lem:lemdiffiterates}}

The difference between the two iterates is bounded by
\begin{align}
\norm[2]{\vtheta^{t+1} - \tilde \vtheta^{t+1}}
&=
\norm[2]{
\vtheta^t - \eta ( \transp{\mJ} \mP \mJ \vtheta^t - \transp{\mJ} \mP \vy)
-
\left(
\tilde \vtheta^t - \eta ( \transp{\mJ} \mJ \tilde \vtheta^t - \transp{\mJ} \vy)
\right)
} \nonumber \\
&=
\norm[2]{ 
(\mI - \eta \transp{\mJ} \mP \mJ)\vtheta^{t} 
-
(\mI - \eta \transp{\mJ} \mJ)\tilde \vtheta^{t}     
- \eta (\transp{\mJ}\vy - \transp{\mJ} \mP \vy) }
\nonumber \\
&=
\norm[2]{ 
(\mI - \eta \transp{\mJ} \mP \mJ)\vtheta^{t}
-
(\mI - \eta \transp{\mJ} \mP \mJ) \tilde \vtheta^{t}
+
(\mI - \eta \transp{\mJ} \mP \mJ) \tilde \vtheta^{t}
-
(\mI - \eta \transp{\mJ} \mJ)\tilde \vtheta^{t}     
- \eta (\transp{\mJ}\vy - \transp{\mJ} \mP \vy) } \nonumber \\
&=
\norm[2]{ 
(\mI - \eta \transp{\mJ} \mP \mJ)(\vtheta^{t} - \tilde \vtheta^{t}) 
+
\eta (\transp{\mJ}\mJ - \transp{\mJ} \mP \mJ ) \tilde \vtheta^{t} 
- \eta (\transp{\mJ}\vy - \transp{\mJ} \mP \vy)
} \nonumber \\
&\leq
\norm{ 
\mI - \eta \transp{\mJ} \mP \mJ
}
\norm[2]{\vtheta^{t} - \tilde \vtheta^{t}} 
+
\eta 
\norm[2]{\transp{\mJ}( \mI -  \mP ) (\mJ \tilde \vtheta^{t} - \vy) } \nonumber \\
&\mystackrel{(i)}{\leq}
\norm[2]{\vtheta^{t} - \tilde \vtheta^{t}}
+
\eta 
\norm[2]{\transp{\mJ}( \mI -  \mP ) (\mJ \tilde \vtheta^{t} - \vy) },
\label{eq:ineqdiffcoef}
\end{align}
where inequality (i) holds for a sufficiently small stepsize, specifically for a stepsize  smaller than $\eta \leq \frac{1}{\sigma_{\max}(\transp{\mJ}\mP \mJ)}$. Note that a sufficiently small stepsize is required for gradient descent to converge. 
We next bound the term on the RHS. Using that 
\begin{align*}
\mI - \mP = 
\begin{bmatrix}
\mI - \transp{\mS} \mS & \mathbf{0} \\
\mathbf{0} & \mathbf{0}
\end{bmatrix},
\end{align*}
and the definition of $\mJ$ and $\vy$ in equation~\eqref{eq:defJ}, we get
\begin{align}
\norm[2]{\transp{\mJ}( \mI -  \mP ) (\mJ \tilde \vtheta^{t} - \vy) }
&=
\norm[2]{\transp{\mJ}_A( \mI -  \transp{\mS}\mS )
\mJ_A ( \tilde \vtheta^{t} - \vtheta_A) }  \nonumber \\ 
&\mystackrel{(i)}{\leq}
8 \frac{\norm[F]{\mJ}}{\sqrt{s}}
\norm[2]{\mJ_A (\tilde \vtheta^{t} - \vtheta_A)} \nonumber \\
&\mystackrel{(ii)}{\leq}
8 \frac{\norm[F]{\mJ_A}}{\sqrt{s}} \norm[2]{\tilde \vr_{t}}.
\label{eq:boundtocont}
\end{align}
Here, inequality (i) holds with probability at least $1 - 4 e^{- \frac{\norm[F]{\mJ_A}^2}{2 \norm{\mJ_A}^2 } }$, as established by the lemma below. The lemma below applies to a setup where the vector $\vz = \mJ_A (\tilde \vtheta^{t} - \vtheta_A)$ is independent of the random matrix $\mS$. The vector $\vz$ is independent of the random matrix $\mS$ since $\tilde \vtheta^{t}$ are the non-sketched gradient iterations.
Moreover, inequality (ii) follows by using that 
$\norm[2]{\mJ_A (\tilde \vtheta^{t} - \vtheta_A)} \leq \norm[2]{\mJ \tilde \vtheta^{t} - \vy} = \norm[2]{\tilde \vr_t}$, where we defined the residual $\tilde \vr_t = \mJ \tilde \vtheta^t - \vy$.

\begin{lemma}
\label{lem:probbound}
For any $\mJ$ and any $\vz \in \reals^n$, and for $\mS \in \reals^{s \times n}$ a random projection matrix with iid $\mc N(0,1/s)$ entries, we have that
\begin{align*}
\PR{
\norm[2]{\transp{\mJ}( \mI -  \transp{\mS}\mS ) \vz }  
\leq 8 \norm[F]{\mJ} \norm[2]{\vz} \frac{1}{\sqrt{s}}
} \geq 1 - 4 e^{- \frac{\norm[F]{\mJ}^2}{2 \norm{\mJ}^2 } }
\end{align*}
\end{lemma}
Application of the bound~\eqref{eq:boundtocont} to inequality~\eqref{eq:ineqdiffcoef}, establishes that 
\begin{align*}
\norm[2]{\vtheta^{t+1} - \tilde \vtheta^{t+1}}
\leq 
\norm[2]{\vtheta^{t} - \tilde \vtheta^{t}}
+ 
8 \frac{\norm[F]{\mJ_A}}{\sqrt{s}} \eta \norm[2]{\tilde \vr_{t}},
\end{align*}
with probability at least $1 - 4 e^{- \frac{\norm[F]{\mJ_A}^2}{2 \norm{\mJ_A}^2 } }$. 
Applying the union bound over $t$ iterations, it follows that, with probability at least $1 - 4t e^{- \frac{\norm[F]{\mJ_A}^2}{2 \norm{\mJ_A}^2 } }$,
\begin{align}
\label{eq:thetadiff}
\norm[2]{\vtheta^{t} - \tilde \vtheta^{t}}
&\leq
8 \frac{\norm[F]{\mJ_A}}{\sqrt{s}}
\eta \sum_{\tau=0}^{t-1}  \norm[2]{\tilde \vr_\tau}.
\end{align}
We next bound the sum of the residuals above. 
Let $\mJ = \mU \mSigma \transp{\mV}$ be the singular value decomposition of $\mJ$, and note that
\begin{align*}
\tilde \vr_t
&=
\mJ\vtheta^t - \vy \\
&= (\mI - \eta \transp{\mJ}{\mJ})^{t} \tilde \vr_0 \\
&= \mU(\mI - \eta \mSigma^2)^t 
\transp{\mU} \tilde \vr_0.
\end{align*}
Let $\mU_{r}$ and $\mU_{n}$ be the singular vectors corresponding to the $r$-leading and the other singular values. 
With this notation, we have that
\[
\norm[2]{\tilde \vr_\tau} 
\leq 
(1-\eta\sigma_{r}^2)^2 \norm[2]{\transp{\mU}_{r} \tilde \vr_0}
+
(1-\eta\sigma_{\min}^2)^2 \norm[2]{\transp{\mU}_{n} \tilde \vr_0}
\]
We therefore can proceed with bounding the RHS of~\eqref{eq:thetadiff} as
\begin{align*}
\norm[2]{\vtheta^{t} - \tilde \vtheta^{t}}
%
%
&\leq 
8 \frac{\norm[F]{\mJ_A}}{\sqrt{s}}
\eta
\sum_{\tau=0}^{t-1} 
(1-\eta\sigma_{r}^2)^\tau \norm[2]{\transp{\mU}_{r} \tilde \vr_0}
+
(1-\eta\sigma_{\min}^2)^2 \norm[2]{\transp{\mU}_{n} \tilde \vr_0}
\\
&=
8 \frac{\norm[F]{\mJ_A}}{\sqrt{s}}
\eta \left(
 \frac{1 - (1-\eta \sigma_{r}^2)^t}{\eta \sigma_{r}^2}
\norm[2]{\transp{\mU}_{r} \tilde \vr_0}
+
\frac{1 - (1-\eta \sigma_{\min}^2)^t}{ \eta \sigma_{\min}^2}
\norm[2]{ \transp{\mU}_{n} \tilde \vr_0}
\right),
\end{align*}
where the last inequality follows from the formula of a geometric series. This concludes the proof of Lemma~\ref{lem:lemdiffiterates}.

\subsection{Proof of Lemma~\ref{lem:probbound}}
It remains to prove Lemma~\ref{lem:probbound}. 
Towards this goal, let $\mP \in \reals^{n\times n}$ be a orthonormal projection onto $\vz$, and let $\mP_\perp\in \reals^{n\times n}$ be a orthonormal projection on the orthogonal complement. 
With this notation, we have 
\begin{align}
\norm[2]{\transp{\mJ}( \mI -  \transp{\mS}\mS ) \vz} 
&= 
\norm[2]{\transp{\mJ} (\mP + \mP_\perp) ( \mI -  \transp{\mS}\mS ) \vz} \nonumber \\
&\mystackrel{(i)}{\leq}
\norm[2]{\transp{\mJ}\mP( \mI -  \transp{\mS}\mS ) \vz}
+ 
\norm[2]{\transp{\mJ} \mP_\perp \transp{\mS}\mS  \vz },
\nonumber \\
&\leq
\norm[2]{\vz} 
\epsilon \norm{\mJ}
+
4 \norm[F]{\mJ} \frac{\norm[2]{\vz}}{\sqrt{s}} \nonumber \\
&\leq
8 \norm[2]{\vz} \norm[F]{\mJ} \frac{1}{\sqrt{s} }, \nonumber
\end{align}
where inequality (i) follows by the triangle inequality, and inequality (ii) holds with probability at least 
$1 - 4e^{- \frac{\norm[F]{\mJ}^2}{2 \norm{\mJ}^2 } }$
using that
\begin{align}
\label{eq:ineq2}
\PR{
\norm[2]{\transp{\mJ}\mP( \mI -  \transp{\mS}\mS ) \vz}
\leq
\norm{\mJ}
\norm[2]{\vz} \epsilon
} \geq 1 - 2e^{- s\frac{\epsilon^2}{12}}
\end{align} 
with the choice of $\epsilon = 4 \frac{1}{\sqrt{s}}\frac{\norm[F]{\mJ}}{ \norm{\mJ} }$, and using that
\begin{align}
\label{eq:ineq1}
\PR{ \norm[2]{\transp{\mJ} \mP_\perp \transp{\mS}\mS  \vz}
\leq 4 \norm[F]{\mJ} \frac{\norm[2]{\vz}}{\sqrt{s}}
} 
\geq 1 - 2e^{- \frac{\norm[F]{\mJ}^2}{2 \norm{\mJ}^2 } }. 
\end{align}

It remains to prove the bounds~\eqref{eq:ineq2} and~\eqref{eq:ineq1}. We start with inequality~\eqref{eq:ineq2}. We have that
\begin{align*}
\norm[2]{\transp{\mJ}\mP( \mI -  \transp{\mS}\mS ) \vz}
&\leq 
\norm{\mJ}
\norm[2]{\mP ( \mI -  \transp{\mS}\mS ) \vz} \\
&=
\norm{\mJ}
\norm[2]{\frac{\vz}{\norm[2]{\vz}} \frac{\transp{\vz}}{\norm[2]{\vz}}  ( \mI -  \transp{\mS}\mS ) \vz} \\
&=
\norm{\mJ}
\frac{1}{\norm[2]{\vz}}
\left| \norm[2]{\vz}^2 - \norm[2]{\mS \vz}^2 \right| \\
&\leq
\norm{\mJ}
\norm[2]{\vz} \epsilon,
\end{align*}
where the last inequality holds with probability at least $1 - 2e^{- s\frac{\epsilon^2}{12}}$ for $\epsilon \in (0,1)$, with a standard concentration inequality for Gaussian matrices. This concludes the proof of the bound~\eqref{eq:ineq2}.

We next prove the bound~\eqref{eq:ineq1}. We need to bound the norm of $\transp{\mJ} \mP_\perp \transp{\mS}\mS  \vz$. 
Note that the terms $\transp{\mJ} \mP_\perp \transp{\mS}$ and $\mS  \vz$ are independent. 
Moreover, $\mS  \vz$ is a Gaussian random vector with iid $\mc N(0,\norm[2]{\vz}^2/s)$ entries. 
We therefore have that
\begin{align}
\norm[2]{\transp{\mJ} \mP_\perp \transp{\mS}\mS  \vz }
=
\frac{\norm[2]{\vz}}{\sqrt{s}}
\norm[2]{\mA \vg} ,
\end{align}
where $\vg$ is a Gaussian vector with iid standard Gaussian entries, independent of $\mS$, and $\mA = \transp{\mJ} \mP_\perp\transp{\mS}$, for notational convenience. 

Recall that for a Gaussian vector $\vg$ with iid standard Gaussian entries and a $L$-Lipschitz function, we have
\begin{align*}
\PR{  f(\vg) - \EX{f(\vg)}  \geq t } \leq e^{- \frac{t^2}{2L^2}}. 
\end{align*}
Using that $f(\vg) = \norm[2]{\mA \vg}$ is $\norm{\mA}$-Lipschitz, 
we get that 
\begin{align*}
\PR{ \norm[2]{\mA \vg} \geq  2 \norm[F]{\mA} }
&= \PR{ \norm[2]{\mA \vg} \geq  \sqrt{ \EX{\norm[2]{\mA\vg}^2} } 
+ \norm[F]{\mA} } \\
&\mystackrel{(i)}{\leq} \PR{ \norm[2]{\mA \vg} \geq   \EX{\norm[2]{\mA \vg}}  
+ \norm[F]{\mA} } \\
&\mystackrel{(ii)}{\leq} 
e^{- \frac{\norm[F]{\mA}^2}{2 \norm{\mA}^2 } }.
\end{align*}
where inequality (i) is by Jensen's inequality (which implies 
$(\EX{\norm[2]{\mA \vg} })^2 \leq \EX{\norm[2]{\mA \vg}^2 }$) 
and inequality (ii) follows by the Gaussian concentration inequality stated above. 
Similarly, we obtain 
\begin{align*}
\PR{ \norm[F]{\mB \mS} \geq  2 \norm[F]{\mB} }
\leq
e^{- \frac{\norm[F]{\mB}^2}{2 \norm{\mB}^2 } }.
\end{align*}
Combining those two inequalities, we get that
\begin{align*}
\norm[2]{\transp{\mJ} \mP_\perp \transp{\mS}\mS  \vz}
=
\norm[2]{\transp{\mJ} \mP_\perp\transp{\mS} \vg} \frac{\norm[2]{\vz}}{\sqrt{s}} 
\leq 2 \norm[F]{\transp{\mJ} \mP_\perp\transp{\mS}} \frac{\norm[2]{\vz}}{\sqrt{s}} 
\leq 4 \norm[F]{\mJ \mP_\perp} \frac{\norm[2]{\vz}}{\sqrt{s}},
\leq 4 \norm[F]{\mJ} \frac{\norm[2]{\vz}}{\sqrt{s}},
\end{align*}
where the first inequality holds with probability at least $1-e^{- \frac{\norm[F]{\mJ}^2}{2 \norm{\mJ}^2 } }$, and the second as well, therefore by the union bound the entire inequality holds with probability at least $1-2e^{- \frac{\norm[F]{\mJ}^2}{2 \norm{\mJ}^2 } }$. This concludes the proof of bound~\eqref{eq:ineq1}.

\subsection{Proof of Theorem~\ref{thm:randproj}, part ii}

Equation~\eqref{eq:ineqdiffcoef} state that
\begin{align}
\norm[2]{\vtheta^{t+1} - \tilde \vtheta^{t+1}}
&\leq
\norm[2]{\vtheta^{t} - \tilde \vtheta^{t}}
+
\eta 
\norm[2]{\transp{\mJ}( \mI -  \mP ) (\mJ \tilde \vtheta^{t} - \vy) }.
\end{align}
Since the matrix $\mJ$ has rank $r$, the residual $(\mJ \tilde \vtheta^{t} - \vy)$ lies in a $(r+1)$-dimensional subspace, for any $\vtheta^{t}$. 
It follows that, with probability at least $1-2e^{-r^2}$, 
\begin{align}
\norm[2]{\transp{\mJ}( \mI -  \mP ) (\mJ \tilde \vtheta^{t} - \vy) }
\leq
\sigma_{\max} \sqrt{ c\frac{r}{d} } \norm[2]{ \mJ \tilde \vtheta^{t} - \vy}.
\end{align}
This probability bound holds for all $t$ simultaneously. Proceeding analogously as in the proof of Lemma~\ref{lem:lemdiffiterates}, we get
 \begin{align*}
\norm[2]{\vtheta^{t} - \tilde \vtheta^{t}}
%
%
&\leq 
\sigma_{\max} \sqrt{ c\frac{r}{d} }
\eta
\sum_{\tau=0}^{t-1} 
(1-\eta\sigma_{r}^2)^2 \norm[2]{\tilde \vr_0}
\\
&=
\sigma_{\max} \sqrt{ c\frac{r}{d} }
\eta 
 \frac{1 - (1-\eta \sigma_{r}^2)^t}{\eta \sigma_{r}^2}
\norm[2]{\tilde \vr_0},
\end{align*}
where the last inequality follows from the formula of a geometric series. This concludes the proof of Theorem~\ref{thm:randproj}, part ii.

\section{Proof of the results in Section~\ref{sec:contlearregression}}

In this section, we prove the two claims we made in Section~\ref{sec:contlearregression}, specifically that equations~\eqref{eq:claim1eq} and \eqref{eq:comperrors} hold with high probability.

\paragraph{Claim 1:}
We first show that if we apply gradient descent for $O(\log(n) / \log(n/d))$ iterations to the loss in~\eqref{eq:lossall}, i.e., to
\begin{align*}
\tilde {\mc L}_{AB}(\vtheta)
=
\mc L_{A}(\vtheta) + \mc L_{B}(\vtheta)
=
\frac{1}{2} \norm[2]{\mJ \vtheta - \vy}^2,
\end{align*}
then the corresponding estimate $\tilde \vtheta_t$ obeys, with high probability,
\begin{align*}
\norm[2]{ \tilde \vtheta_t - \vtheta^\ast}
\leq O\left( \sqrt{\frac{d}{n}} ( \norm[2]{\vtheta_A} +  \norm[2]{\vtheta_B} + \sigma)  \right).
\end{align*}
To establish this claim, we first note that the extreme singular values of a Gaussian matrix satisfy, for $t>0$,~\cite[Equation 2.3]{rudelson_NonasymptoticTheoryRandom_2010} 
\begin{align}
\PR{\sqrt{n} - \sqrt{d} - t  \leq \sigma_{\min} \leq \sigma_{\max} \leq \sqrt{n} + \sqrt{d} + t } 
\geq 1 - 2 e^{-t^2/2}.
\end{align}
With $t = \sqrt{d}$, we get
\begin{align}
\label{eq:singvalsgaussian}
\PR{\sqrt{n} - 2\sqrt{d}   \leq \sigma_{\min} \leq \sigma_{\max} \leq \sqrt{n} + 2\sqrt{d} } 
\geq 1 - 2 e^{-d^2/2},
\end{align}
which we use below to establish the result. 

Next, note that the minimizer of $\tilde {\mc L}_{AB}(\vtheta)$ is given by 
\begin{align*}
\hat \vtheta 
&= \inv{(\transp{\mJ} \mJ)} \transp{\mJ}
\begin{bmatrix}
\mJ_A \inv{(\transp{\mJ}_A \mJ_A)} \transp{\mJ}_A \vy_A \\
\vy_B
\end{bmatrix} \\
&= \inv{(\transp{\mJ} \mJ)} 
(\transp{\mJ}_A \mJ_A \inv{(\transp{\mJ}_A \mJ_A)} \transp{\mJ}_A \vy_A + \transp{\mJ}_B \vy_B) \\
&= \inv{(\transp{\mJ} \mJ)} (\transp{\mJ}_A \vy_A + \transp{\mJ}_B \vy_B).
\end{align*}
It follows that
\begin{align}
\norm[2]{\hat \vtheta - \vtheta^\ast}
&= \norm[2]{\hat \vtheta - \vtheta_A - \vtheta_B} \nonumber \\
&= 
\norm[2]{
 \inv{(\transp{\mJ} \mJ)} (\transp{\mJ}_A \mJ_A \vtheta_A + \vz_A) + \transp{\mJ}_B(\mJ_B \vtheta_B + \vz_B)
 - \vtheta_A - \vtheta_B} \nonumber \\
&\leq
\norm[2]{\inv{(\transp{\mJ} \mJ)} (\transp{\mJ}_A \mJ_A \vtheta_A + \vz_A)  - \vtheta_A}
+
\norm[2]{
\inv{(\transp{\mJ} \mJ)} \transp{\mJ}_B(\mJ_B \vtheta_B + \vz_B) - \vtheta_B} \nonumber \\
&\leq
c \sqrt{\frac{d}{n}}
\left(
\norm[2]{\vtheta_A}
+
\norm[2]{\vtheta_B}
+ 
\sigma
\right),
\label{eq:bhatstar}
\end{align}
where the last inequality holds with probability at least $1- 8e^{-d^2/2}$, and follows from 
\begin{align*}
\norm[2]{\inv{(\transp{\mJ} \mJ)} \transp{\mJ}_A (\mJ_A \vtheta_A + \vz_A)  - \vtheta_A} 
&\leq
\norm{\mI - \inv{(\transp{\mJ} \mJ)} (\transp{\mJ}_A \mJ_A)}
\norm[2]{\vtheta_A}
+
\norm{\inv{(\transp{\mJ} \mJ)} } \norm[2]{\transp{\mJ}_A \vz} \\
&\leq
c \sqrt{\frac{d}{n}}
\norm[2]{\vtheta_A}
+
c \sqrt{\frac{d}{n}} \sigma,
\end{align*}
where we used inequality~\eqref{eq:singvalsgaussian}, and where $c$ is a numerical constant. Specifically, here we used that, with probability at least $1- 8e^{-d^2/2}$,
\begin{align*}
\norm{\mI - \inv{(\transp{\mJ} \mJ)} (\transp{\mJ}_A \mJ_A)}
&\leq 
\left| 1 -  \frac{ (\sqrt{n} +2\sqrt{d})^2 }{ (\sqrt{n} -2\sqrt{d})^2 } \right|
= 
\left| \frac{ 4\sqrt{n}\sqrt{d} }{ n + 2d - 2 \sqrt{n} \sqrt{d} } \right| 
\leq \frac{4 \sqrt{d}}{\sqrt{n}},
\end{align*}
and that 
\begin{align*}
\norm{\inv{(\transp{\mJ} \mJ)} } \norm[2]{\transp{\mJ}_A \vz}
\leq
\frac{1}{ (\sqrt{n} - 2\sqrt{d})^2 } c \sqrt{n d} \sigma
\leq
c \frac{\sqrt{d}}{ \sqrt{n} },
\end{align*}
again with high probability. 
Here, we used that the entries of $\mJ_A \vz \in \reals^d$, conditioned on $\vz$,  are iid Gaussian with norm $\norm[2]{\vz}$, and that $\norm[2]{\vz}$ concentrates around $\sigma \sqrt{n}$. 

Next, let $\mJ = \mU \mSigma \transp{\mV}$ be the singular value decomposition of $\mJ$. The gradient descent iterations with stepsize $\eta$ starting at $\tilde \vtheta = \mathbf{0}$ are
\begin{align*}
\tilde \vtheta^t = \mV \mD^t \transp{\mU} \vy,
\end{align*}
where $\mD^t$ is a diagonal matrix with $i$-th entry given by $\frac{1 - (1-\eta \sigma_i^2)^t}{\sigma_i}$. 
With sufficiently small stepsize, gradient descent converges to the minimizer of the loss, i.e., $\hat \vtheta = \tilde \vtheta^\infty$. 
Thus,
\begin{align*}
\norm[2]{\tilde \vtheta^t - \hat \vtheta} 
&= 
\norm[2]{ \sum_{i=1}^d \vv_i (1 - \eta \sigma_i^2)^t \innerprod{\vu_i}{\vy} } \\
&\leq
\norm[2]{\vy} (1 - \eta \sigma_{\min}^2)^t \\
&\leq
c \sqrt{n} (\norm[2]{\vtheta_A} + \norm[2]{\vtheta_B} + \sigma ) \left( \frac{ 8 \sqrt{d} }{ \sqrt{n} } \right)^t,
\end{align*}
where again the last inequality holds with high probability. 
Choosing the stepsize as $1/\sigma_{\max}^2$, we get with 
\[
1 - \frac{\sigma_{\min}^2}{\sigma_{\max}^2}
\leq
1 - \frac{ (\sqrt{n} + 2\sqrt{d})^2 }{ (\sqrt{n} - 2\sqrt{d})^2 }
=
\frac{ 8\sqrt{n}\sqrt{d} }{ n + d  - 8\sqrt{n}\sqrt{d} } \leq \frac{ 8 \sqrt{d} }{ \sqrt{n} }
\]
that 
\begin{align}
\norm[2]{\tilde \vtheta^t - \hat \vtheta} 
&\leq
c (\norm[2]{\vtheta_A} + \norm[2]{\vtheta_B} + \sigma ) \left( \frac{ 8
\sqrt{d} }{ \sqrt{n} } \right)^t \nonumber \\
&\leq
c (\norm[2]{\vtheta_A} + \norm[2]{\vtheta_B} + \sigma ) 
\frac{ \sqrt{d} }{ \sqrt{n} },
\label{eq:gaeraion}
\end{align}
provided that $t \geq \frac{ 2\log(n) }{ \log(n) - \log(64d) }$. 
It follows that for $t \geq O(\log(n) / \log(n/d))$
\[
\norm[2]{ \tilde \vtheta_t - \vtheta^\ast }
\leq
\norm[2]{ \tilde \vtheta_t - \hat \vtheta } + \norm[2]{ \hat \vtheta - \vtheta^\ast }
\leq
c (\norm[2]{\vtheta_A} + \norm[2]{\vtheta_B} + \sigma ) 
\frac{ \sqrt{d} }{ \sqrt{n} },
\]
where we used the previously established inequalities~\eqref{eq:bhatstar} and~\eqref{eq:gaeraion}. 
This establishes equation~\eqref{eq:claim1eq} as claimed.

\paragraph{Claim 2:}
The second claim we made in Section~\ref{sec:contlearregression} is that (cf.~equation~\eqref{eq:comperrors}):
\begin{align}
\label{eq:comperrorsagain}
\norm[2]{\vtheta_t - \vtheta^\ast}
&\leq O\left( \left( \sqrt{\frac{d}{n}} + \sqrt{ \frac{d}{s}} \right)  ( \norm[2]{\vtheta_A} +  \norm[2]{\vtheta_B} + \sigma)  \right).
\end{align}
This claim follows directly from combining equation~\eqref{eq:claim1eq} with Theorem~\ref{thm:randproj}. Specifically, note that for the setup in Section~\ref{sec:contlearregression}, Theorem~\ref{thm:randproj} gives, with $r = d$, that
\begin{align*}
\norm[2]{\vtheta_{t} - \tilde \vtheta_{t}} 
&\leq 
8 \frac{ \norm[F]{\mJ_A} }{\sqrt{s} \sigma_{{\min}} } \frac{1}{ \sigma_{\min} }
\norm[2]{ \vy } \\
&\leq
 \frac{ \sqrt{dn} }{\sqrt{s} \sqrt{n} } \frac{1}{ \sqrt{n} }
 \sqrt{n} (\norm[2]{\vtheta_A} + \norm[2]{\vtheta_B} + \sigma ) \\
&=
 \frac{ \sqrt{d} }{\sqrt{s} } (\norm[2]{\vtheta_A} + \norm[2]{\vtheta_B} + \sigma ) 
\end{align*}
holds with high probability, provided that $n \geq O(d)$. Specifically, we used,
$\norm[F]{\mJ_A} \leq O(\sqrt{ nd })$, $\sigma_{{\min}} \geq c \sqrt{n}$ (which holds for $n \geq O(d)$), and as established before, $\norm[2]{\vy} \leq c \sqrt{n} (\norm[2]{\vtheta_A} + \norm[2]{\vtheta_B} + \sigma )$. All three inequalities hold with high probability.
Application of this inequality to
$\norm[2]{\vtheta_{t} - \vtheta^\ast}
\leq
\norm[2]{\vtheta_{t} - \tilde \vtheta_{t}} 
+
\norm[2]{\tilde \vtheta_{t} - \vtheta^\ast} 
$ establishes the bound~\eqref{eq:comperrorsagain}.


\section{Proof of Theorem~\ref{thm:ntk}: Guarantees for two-layer neural networks}

Let $\mJ_A, \mJ_B \in \reals^{n \times dk}$ be the Jacobians of the network's predictions for the training sets of task $A$ and task $B$ at initialization. At initialization, each entry of the the weight matrix $\mTheta$ is initialized by drawing a zero-mean Gaussian with variance $\omega^2$. 

The Jacobians depend on the network's parameter, but if the network is sufficiently wide, the Jacobians change very little during gradient descent iterations, and if the network is in the NTK regime (and thus is infinitely wide), the Jacobians are constant and do not change across gradient descent iterations. 
To simplify exposition, we work in the NTK regime where the Jacobians are constant throughout gradient descent iterations. We comment on changes that can be made to establish a result where the network is wide, but not infinitely wide, and thus the Jacobians vary little.

We provide a bound on the composite risk in equation~\eqref{eq:riskntk} by decomposing the risk into the empirical risk and generalization errors of the two tasks
\begin{align}
R(f)
&= \frac{1}{2} R_A(f) + \frac{1}{2} R_B(f) \nonumber \\
&= \frac{1}{2} \hat R_A(f) + \frac{1}{2} \hat R_B(f) 
+ \frac{1}{2}( R_A(f) - \hat R_A(f) ) 
+ \frac{1}{2}( R_B(f) - \hat R_B(f) ), 
\label{eq:overallrisk}
\end{align}
and by bounding the empirical risks and generalization errors separately. Here, the empirical risk of task $T$ is $\hat R_T(\vtheta) = \sum_{i=1}^n \loss(f_\vtheta(\vx_{T,i}) ,y_{T,i} )$, where the $(\vx_{T,i},y_{T,i})$'s are the training data pertaining to task $T = \{A,B\}$. 


\paragraph{Bounding the empirical risk:}
For bounding the empirical risk, we rely on the following lemma which ensures that the norm of the residual of the sketched problem, which is square-root of the empirical risk, is close to the residual of the norm of the original, non-sketched problem. We use the same notation as in the previous section, specifically we define: 
\begin{align*}
\mJ 
= 
\begin{bmatrix}
\mJ_A \\ \mJ_B
\end{bmatrix}
\in \reals^{2n \times dk}
,
\quad
\mP = \begin{bmatrix}
\transp{\mS} \mS & \mathbf{0} \\ \mathbf{0} & \mI
\end{bmatrix},
\quad
\vy
=
\begin{bmatrix}
\mJ_A \vtheta_A \\ \vy_B
\end{bmatrix}.
\end{align*}

\begin{lemma}
\label{lem:residualsdiff}
Let $\vr_\iter = \mP^{1/2}\mJ \vtheta^t - \vy$ be the residual associated with the sketched least-squares problem and let $\tilde \vr_\iter = \mJ \vtheta^t - \vy$ be the residual associated with the original least-squares problem. With probability at least $1- 4 t \exp( - \frac{\norm[F]{\mJ_A}^2}{2 \norm{\mJ_A}^2 } )$ over the random sketch $\mS \in \reals^{s\times n}$ the residuals are close:
\begin{align}
\label{eq:residualsclose}
\norm[2]{\vr^{\iter} - \tilde \vr^{t}}
\leq \frac{ \norm[F]{\mJ \transp{\mJ}} }{ \sqrt{s} } \frac{1}{\alpha^2}.
\end{align}
\end{lemma}

With this lemma in place, we note that, after $t$ iterations of gradient descent, we have
\begin{align}
\sqrt{ 
\sum_{i=1}^n (f_{\vtheta_t}(\vx_{A,i}) - y_{A,i} )^2 
+
\sum_{i=1}^n (f_{\vtheta_t}(\vx_{B,i}) - y_{B,i} )^2 
}
&\mystackrel{(i)}{=} \frac{1}{\sqrt{n}} \norm[2]{\mJ \vtheta_\iter - \vy} = \frac{1}{\sqrt{n}} \norm[2]{\vr_\iter} \nonumber \\
&\leq \frac{1}{\sqrt{n}} \norm[2]{ \tilde \vr_\iter } + \frac{1}{\sqrt{n}} \norm[2]{\vr_\iter - \tilde \vr_\iter} \nonumber \\
&\mystackrel{(ii)}{\leq}
\sqrt{ \frac{1}{n}\sum_{i=1}^{2n}  \innerprod{\vu_i}{\vy}^2 (1-\eta \sigma_i^2)^{2t} }  
+ 
\frac{1}{\sqrt{n}} \frac{ \norm[F]{ \mK} }{ \sqrt{s} } \frac{1}{\alpha^2}.
\label{eq:empriskbound1}
\end{align}
Here, equation (i) holds if we are in the NTK regime and thus the predictions of the network are its Jacobian at initialization times the model parameter. For finite-width networks the equality holds up to an error that goes to zero as the network's width tends to infinity. 
Inequality (ii) holds with probability at least $1- 4 t \exp( - \frac{\norm[F]{\mJ_A}^2}{2 \norm{\mJ_A}^2 } )$ by Lemma~\ref{lem:residualsdiff}, and by using that $\transp{\mJ} \mJ = \mK$. 
With this bound, we obtain
\begin{align*}
\emprisk_A({\vtheta_\iter})
&= 
\frac{1}{n} \sum_{i=1}^n \loss( f_{\vtheta_\iter} (\vx_{A,i}), y_{A,i}  ) \\
&\mystackrel{(i)}{\leq}
\frac{1}{n} \sum_{i=1}^n | f_{\vtheta_\iter} (\vx_{A,i}) - y_{A,i}  | \\
&\leq
\sqrt{
\frac{1}{n} \sum_{i=1}^n
(f_{\vtheta_\iter}(\vx_{A,i}) - y_{A,i})^2
}
\\
&\mystackrel{(ii)}{\leq}
\sqrt{
\frac{1}{n}
\sum_{i=1}^{2n} \innerprod{\vu_i}{\vy}^2 (1 - \eta \sigma_i^2)^{2\iter}
}
+ \frac{1}{\sqrt{n}} \frac{ \norm[F]{ \mK} }{ \sqrt{s} } \frac{1}{\alpha^2}.
\end{align*}
Here the sum in the first three equations is over trainings examples from task $A$, and 
equation (i) follows from
$\loss(z,y) = \loss(z,y) - \loss(y,y) \leq |z- y|$ 
because the loss is $1$-Lipschitz. 
Equation~(ii) follows from equation~\eqref{eq:empriskbound1} above.

The same bound holds for the risk of task $B$, $\hat R_B(\vtheta_t)$. 

\paragraph{Bonding the generalization error:}
We bound the generalization error for task $A$ and task $B$ separately. The derivations for each bound is the same, so we detail the derivations for task $A$ only. We bound the generalization error of task $A$ by bounding the Rademacher complexity of the class of functions that gradient descent can reach with $\iter$ iterations. This is a common proof technique, see for example the papers~\citep{mohri_FoundationsMachineLearning_2012,arora_FineGrainedAnalysisOptimization_2019,heckel_EarlyStoppingDeep_2021}. 

Let $\mc F$ be a set of functions mapping a $d$-dimensional feature vector to a real number, and let $\epsilon_1,\ldots,\epsilon_n$ be iid Rademacher random variables. A Rademacher random variable is chosen uniformly from $\{-1,1\}$.  
The \emph{empirical Rademacher complexity} of the function class $\setF$ is defined as 
\begin{align*}
\mc R_{\setD}(\setF)
=
\frac{1}{n}
\EX[\vepsilon]{ \sup_{f \in \setF} \sum_{i=1}^n \epsilon_i f(\vx_i) },
\end{align*}
where $\setD =\{(\vx_1,y_1),\ldots, (\vx_n,y_n)\}$ is a training set containing $n$ examples drawn iid from the distribution pertaining to task $A$. 
The following theorem bounds the generalization error uniformly over all functions in the class $\setF$ with the empirical Radermacher complexity of the function class $\setF$.  

\begin{theorem}[{ \cite[Thm.~3.1]{mohri_FoundationsMachineLearning_2012} }]
Assume the loss $\loss(\cdot,\cdot)$ is bounded in $[0,1]$ and $1$-Lipschitz in its first argument. 
With probability at least $1-\delta$ over the set $\setD$ consisting of $n$-many iid examples the generalization error is bounded by
\begin{align*}
\sup_{f \in \setF}
\risk(f) - \emprisk(f)
\leq
2 \mc R_{\setD}(\setF) 
+ 3 \sqrt{ \frac{\log(2/\delta)}{2n} }.
\end{align*}
\end{theorem}

We consider the class of two-layer neural networks of the form as in equation~\eqref{eq:networkmodel} with weights close to the random initialization $\mTheta_0$, defined as:
\begin{align}
\label{eq:defclassnetworks}
\setF_{Q,M}
=
\left\{
f_{\mTheta}
\colon 
\norm[F]{\mTheta - \mTheta_0} \leq  Q, 
\norm[2]{\vtheta_r - \vtheta_{0,r}} \leq \omega M
\right\}, 
\end{align}
where $\vtheta_r$ is the $r$-th column of the weight matrix $\mTheta \in \reals^{d\times k}$. 

The Rademacher complexity of this class of functions is bounded in the following lemma, which is a version of Lemma 5.4 in the paper~\citep{arora_FineGrainedAnalysisOptimization_2019} and a version of Lemma 4 in the paper~\citep{heckel_EarlyStoppingDeep_2021}. 

\begin{lemma}
\label{lem:lemrademacher}
Let $\mTheta_0$ be drawn from a Gaussian distribution with $\mc N(0,\omega^2)$ entries, and suppose half of the entries of $\vv_0$ are $1$ and the other half is $-1$. 
Assume the examples $(\vx_i,y_i)$ are drawn iid from some distribution with $\norm[2]{\vx_i} =1$ and $|y_i| \leq  1$. 
With probability at least $1-\delta$ over the random training set, the empirical Rademacher complexity of $\setF_{Q,M}$ is, simultaneously for all $Q$, bounded by
\begin{align}
\mc R_\setD(\setF_{Q,M})
&\leq
\frac{Q}{\sqrt{n}} 
+
4 \omega M \left( \sqrt{k} M + \sqrt{\log(2/\delta)/2} \right).
%
\end{align}
\end{lemma}

We set $M = O(\frac{\xi}{\alpha} k^{-1/4})$, where $\xi$ is an error tolerance parameter that goes to zero . With this choice,
the term on the right hand side above is bounded by
\[ 
\omega (4M^2 \sqrt{k} + 4M \sqrt{\log(2/\delta)/2})
\leq
O(\xi/\alpha),
\]
where we used $\omega \leq 1$ and $\frac{\sqrt{\log(2/\delta)/2}}{k^{1/4}} \leq 1$ by the network being sufficiently wide. Recall that we consider the regime where $k\to \infty$, so this condition is satisfied. 

Let $Q_i = i $ for $i = 1, 2,\ldots$. Simultaneously for all $i$, by the lemma above, for this choice of $M$, the function class $\setF_{Q_i,M}$ has Rademacher complexity bounded by 
\begin{align}
\mc R_\setD(\setF_{Q_i,M})
\leq
\frac{Q_i}{\sqrt{n}}
+
O(\xi/\alpha).
\end{align}
We next choose the radius $Q$ as 
$
Q = 
\sqrt{
\sum_{i=1}^{2n}
\left(
\innerprod{\vu_{i} }{\vy}
\frac{1 - (1 - \stepsize \sigma^2_{i})^\iter}{\sigma_{i}}
\right)^2
}
+
5 \frac{\norm[F]{\mJ_A}}{\sqrt{s}}
\frac{1}{\alpha^2}
\sqrt{n}
+
\frac{\xi}{\alpha} \sqrt{n},
$
where $\xi$ is an approximation parameter that we can choose arbitrarily small for $k\to \infty$. 

This choice is motivated as follows. We have that 
\begin{align*}
\norm[2]{\vtheta_t - \vtheta_0}
&\leq
\norm[2]{\tilde \vtheta_t - \vtheta_0}
+
\norm[2]{\vtheta_t - \tilde \vtheta_t} \\
&\leq
\sqrt{
\sum_{i=1}^{2n}
\left(
\innerprod{\vu_{i} }{\vy}
\frac{1 - (1 - \stepsize \sigma^2_{i})^\iter}{\sigma_{i}}
\right)^2
}
+
5 \frac{\norm[F]{\mJ_A}}{\sqrt{s}}
\frac{1}{\alpha^2}
\norm[2]{\vy},
\end{align*}
where the last inequality holds by Lemma~\ref{lem:lemdiffiterates} with probability at least $1- 4 t \exp( - \frac{\norm[F]{\mJ_A}^2}{2 \norm{\mJ_A}^2 } )$. The extra term $\frac{\xi}{\alpha} \sqrt{n}$ is due to the error of the Jacobian varying slightly over iterations, and goes to zero as the width $k\to \infty$. 

Let $i^\ast$ be the smallest integer such that $Q \leq Q_{i^\ast}$, so that $Q_{i^\ast} \leq Q+1$. 
We have that $i^\ast \leq O(\sqrt{n/\alpha})$ and
\begin{align}
\mc R_\setD(\setF_{Q_{i^\ast},M})
&\leq
\frac{(Q+1)}{\sqrt{n}}
+
O(\xi/\alpha) \nonumber \\
&\leq
\sqrt{
\frac{1}{n}
\sum_{i=1}^{2n}
\left(
\innerprod{\vu_{i}}{\vy}
\frac{1 - (1 - \stepsize \sigma^2_{i})^\iter}{\sigma_{i}}
\right)^2
}
+
\frac{1}{\sqrt{n}}
+
5 \frac{ \norm[F]{\mJ_A} }{ \sqrt{s} } \frac{1}{\alpha^2} 
+
O(\xi/\alpha).
\end{align}
Next, from a union bound over the finite set of integers $i= 1,\ldots, i^\ast$, we obtain
\begin{align}
\max_{i= 1,\ldots, i^\ast}
\sup_{f \in \setF_{Q_i,M}}
\risk_A(f) - \emprisk_A(f)
&\leq
\sqrt{
\frac{1}{n}
\sum_{i=1}^{2n}
\left(
\innerprod{\vu_{i}}{\vy}
\frac{1 - (1 - \stepsize \sigma^2_{i})^\iter}{\sigma_{i}}
\right)^2
}
+
\frac{1}{\sqrt{n}}
+
5 \frac{ \norm[F]{\mJ_A} }{ \sqrt{s} } \frac{1}{\alpha^2} 
+
O(\xi/\alpha) \nonumber \\
&\leq
\sqrt{
\frac{1}{n}
\transp{\vy} \inv{(\transp{\mJ}\mJ)} \vy
}
+
5 \frac{ \norm[F]{\mJ_A} }{ \sqrt{s} } \frac{1}{\alpha^2} 
+
\frac{1}{\sqrt{n}}
+
O(\xi/\alpha).
\nonumber 
\end{align}

\paragraph{Assembling the bounds:}
Combining the bounds on the empirical risk and on the generalization errors by inserting them in the right-hand-side of equation~\eqref{eq:overallrisk} yields
\begin{align*}
\risk(f_{\vtheta_\iter})
&\leq
\sqrt{ \frac{1}{n}\sum_{i=1}^{2n}  \innerprod{\vu_i}{\vy}^2 (1-\eta \sigma_i^2)^{2t} }  
+ 
\frac{1}{\sqrt{n}} \frac{ \norm[F]{ \mK} }{ \sqrt{s} } \frac{1}{\alpha^2} \\
&+
2\sqrt{
\frac{1}{n}
\transp{\vy} \inv{\mK} \vy
}
+
10 \frac{\norm[F]{\mJ_A} }{ \sqrt{s} } \frac{1}{\alpha^2}
+
\frac{2}{\sqrt{n}}
+
O(\xi/\alpha) 
\\
&\leq
2\sqrt{
\frac{1}{n}
\transp{\vy} \inv{\mK} \vy
}  
+
\frac{3}{\sqrt{n}}  
+
\frac{1}{\sqrt{s} \alpha^2}
\left(
10 \norm[F]{\mJ_A}
+
\frac{\norm[F]{ \mK} }{ \sqrt{n} } 
\right),
\end{align*}
where we upper-bounded the first term by using the assumption $t \geq \frac{\log(1 - \eta \alpha)}{ \log(1/n) }$, and where we also again used the assumption that the network is infinitely wide and thus $\xi \to 0$. 
This concludes the proof of the theorem.


\subsection{Proof of Lemma~\ref{lem:residualsdiff}}

With similar steps as in equation~\eqref{eq:ineqdiffcoef} we get 
\begin{align}
\norm[2]{\vr^{\iter + 1} - \tilde \vr^{t+1}}
&=
\norm[2]{ (\mI - \eta \mJ \transp{\mJ} \mP ) (\vr^{\iter} - \tilde \vr^{t}) }
+ \eta \norm[2]{ \mJ \transp{\mJ} (\mI - \mP) \tilde \vr^\iter } \nonumber \\
&\leq
\norm[2]{ \vr^{\iter} - \tilde \vr^{t}}
+ \eta \norm[2]{ \mJ \transp{\mJ} (\mI - \mP) \tilde \vr^\iter } \nonumber \\
&\leq
\norm[2]{ \vr^{\iter} - \tilde \vr^{t}}
+ \eta \frac{ \norm[F]{\mJ \transp{\mJ}} }{ \sqrt{s} } \norm[2]{\tilde \vr^\iter}, \nonumber
\end{align}
where the last inequality holds with probability at least $1 - 4 e^{- \frac{\norm[F]{\mJ_A}^2}{2 \norm{\mJ_A}^2 } }$ as established by Lemma~\ref{lem:probbound}. 
Applying this inequality recursively, we obtain, by the union bound, that with probability at least $1 - 4 t e^{- \frac{\norm[F]{\mJ_A}^2}{2 \norm{\mJ_A}^2 } }$
\begin{align*}
\norm[2]{\vr^{\iter} - \tilde \vr^{t}}
\leq \eta \frac{ \norm[F]{\mJ \transp{\mJ}} }{ \sqrt{s} }
\sum_{i=0}^{n-1} \norm[2]{\tilde \vr^\iter} \\
\leq \frac{ \norm[F]{\mJ \transp{\mJ}} }{ \sqrt{s} } \frac{1}{\alpha^2}.
\end{align*}
This concludes the proof.

  
\section{Proof of the results in Section~\ref{sec:EWCtheory}}

\subsection{Proof of Theorem~\ref{thm:cansucceed}}

The population loss for task $T$ is
\begin{align*}
\mc L_T(\vtheta)
&=
\EX[(\vx,y)\sim P_T]{( \innerprod{\vx}{\vtheta} - y )^2} \\
&= 
(\innerprod{\vtheta}{\vmu_T} - 1)^2 + \sigma^2 \norm[2]{\vtheta}^2.
\end{align*}

Thus training on task $A$ yields
\begin{align*}
\vtheta_A 
= \inv{ (\sigma^2\mI + \vmu_A \transp{\vmu}_A ) } \vmu_A 
= \frac{1}{1 + \sigma^2 } \vmu_A,
\end{align*}
where the second equality follows from the Sherman-Morrison-Woodbury formula. 

Now consider the loss associated with EWC, given by
\begin{align*}
\mc L_{AB}(\vtheta) 
=
\EX[B]{( \innerprod{\vx}{\vtheta} - y )^2} + \lambda \transp{(\vtheta - \vtheta_A)} \mD_A  (\vtheta - \vtheta_A),
\end{align*}
where $\mD_A$ is the diagonal of the Hessian (or Jacobian outer product) of the loss associated with task $A$, i.e., $\mD_A = \diag( \sigma^2 \mI + \vmu_A \transp{\vmu}_A)$. 
The minimizer of $\mc L_{AB}(\vtheta)$ denoted $\vtheta_{AB}^{EWC}$, is, again by using the Sherman-Morrison formula 
\begin{align*}
\vtheta_{AB}^{EWC}
&= \inv{ (\sigma^2\mI + \vmu_B \transp{\vmu}_B + \lambda \mD_A ) }(\vmu_B + \lambda \mD_A \vmu_A) \\
&= \inv{ ( \mD + \vmu_B \transp{\vmu}_B ) }(\vmu_B + \lambda \mD_A \vmu_A) \\
&=
\left( \inv{\mD} - \frac{1}{1 + \transp{\vmu}_B \inv{\mD} \vmu_B } \inv{\mD} \vmu_B \transp{\vmu}_B \inv{\mD}   \right) 
(\vmu_B + \lambda \mD_A \vmu_A) \\
&=  
\frac{1}{1 + \transp{\vmu}_B \inv{\mD} \vmu_B  }  \inv{\mD} \vmu_B  
+
\lambda 
\left( \mI - \frac{1}{1 + \transp{\vmu}_B \inv{\mD} \vmu_B  } \inv{\mD} \vmu_B \transp{\vmu}_B  \right)
\inv{\mD} \mD_A \vmu_A,
\end{align*}
where we defined $\mD = \lambda \mD_A + \sigma^2 \mI$ for notational convenience. 
Next, we use the assumption that all entries of $\vmu_T$ have absolute value $\sqrt{1/d}$. 
This assumption implies that $\mD_A = (\sigma^2 + 1/d)\mI$ 
and $\mD = q\mI$ with $q=\sigma^2 + \lambda (\sigma^2 + 1/d)$, and yields
\begin{align*}
\vtheta_{AB}^{EWC} 
&=
\frac{1}{1+q} \vmu_B 
+ \frac{\lambda(\sigma^2+\frac{1}{d})}{ q } \vmu_A 
-
\frac{\lambda(\sigma^2+\frac{1}{d})}{(1+q)q} \vmu_B \innerprod{\vmu_A}{\vmu_B} \\
&=
\vmu_B \frac{1}{1+q} 
\left(
1 - 
\frac{\lambda(\sigma^2+1/d)}{q} \innerprod{\vmu_A}{\vmu_B}
\right) 
+
\vmu_A 
\frac{\lambda(\sigma^2+\frac{1}{d})}{ \lambda(\sigma^2 + \frac{1}{d} + \sigma^2 }
\end{align*}
Next, note that the term associated with $\mu_A$ is increasing from $0$ to $1$ in $\lambda$. 
Contrary, the term associated with $\mu_B$ is decreasing in $\lambda$. Thus, $\lambda$ interpolates between linear combinations of $\vmu_A$ and $\vmu_B$ and therefore there exists a regularization parameter $\lambda$ such that $\vtheta_{AB}^{EWC}$ points in the same direction as the optimal parameter $\vtheta_{AB}^\ast = \vmu_A + \vmu_B$, and is therefore Bayes optimal. This concludes the first part of the theorem. The proof of the second part, given below, is analogous.

\subsection{Proof of Theorem~\ref{thm:cansucceed}, part two}

The proof is analogous to the proof of part one of the theorem. 
The loss associated with L2 regularization is given by
\begin{align*}
\mc L_{AB}(\vtheta) 
=
\EX[B]{( \innerprod{\vx}{\vtheta} - y )^2} + \lambda \transp{(\vtheta - \vtheta_A)} (\vtheta - \vtheta_A).
\end{align*}
Application of the Sherman-Morrison formula yields
\begin{align*}
\vtheta_{AB}^{L2}
&= \inv{ (\sigma^2\mI + \vmu_B \transp{\vmu}_B + \lambda \mI ) }(\vmu_B + \lambda \vmu_A) \\
&=
\left(
\frac{1}{\sigma^2 + \lambda} \mI - \frac{1}{1 + 1/(\sigma^2+\lambda)}
\frac{1}{(\sigma^2+\lambda)^2}
\vmu_B \transp{\vmu}_B
\right)
(\vmu_B + \lambda \vmu_A) \\
&=
\frac{1}{\sigma^2 + \lambda}
\left( \mI - \frac{1}{1 + \sigma^2+\lambda}
\vmu_B \transp{\vmu}_B
\right)
(\vmu_B + \lambda \vmu_A) \\
&=
\frac{1}{\sigma^2 + \lambda}
\left(  
\frac{\sigma^2 + \lambda}{1 + \sigma^2+\lambda} \vmu_B 
+\lambda \vmu_A
- \frac{\lambda \innerprod{\vmu_A}{\vmu_B}}{1 + \sigma^2+\lambda}  \vmu_B
\right) \\
&=
\frac{1}{\sigma^2 + \lambda}
\left(  
\frac{\sigma^2 + \lambda - \lambda \innerprod{\vmu_A}{\vmu_B} }{1 + \sigma^2+\lambda} \vmu_B 
+
\lambda \vmu_A
\right).
\end{align*}
The term in front of $\vmu_B$ is decreasing in $\lambda$ and varies from $\frac{\sigma^2}{1+\sigma^2}$ to $0$ as $\lambda$ increases. The term associated with $\vmu_A$ varies from $0$ to $1$. Thus, again, there is a parameter $\lambda$ such the solution is optimal.


\subsection{Proof of Theorem~\ref{thm:canfail}}

To prove this statement, it is sufficient to construct a problem instance consisting of task means $\vm_A$ and $\vmu_B$ and a variance $\sigma^2$ for which the risk of $\vtheta_{AB}^{EWC}(\lambda)$ is large for all $\lambda$. Such an problem instance is 
$\mu_A = [1,-0.8,0.8]$, $\mu_B = [-1,0.5,-0.8]$, and $\sigma$ sufficiently small, as illustrated in the code supplement. 
The proof of the second part is analogous, by constructing a similar problem instance for L2.

\end{document}